\documentclass[preprint,journal]{vgtc}
\graphicspath{{figures/}{pictures/}{images/}{./}}

\usepackage{times}

\usepackage{amsmath}
\usepackage{amssymb}
\usepackage{ragged2e}
\usepackage{tabularx}
\usepackage{booktabs}
\usepackage{array}
\usepackage[table]{xcolor}
\usepackage{pifont}
\usepackage{utfsym}
\DeclareMathAlphabet{\mathcal}{OMS}{cmsy}{m}{n}
\newcommand{\etal}{\textit{et al.}}
\usepackage{mathptmx}
\usepackage[utf8]{inputenc}
\vgtccategory{Research}

\title{Setup-Independent Full Projector Compensation}
\supptitle{Setup-Independent Full Projector Compensation}

\author{%
  \authororcid{Haibo Li}{0009-0006-9201-104X},
  \authororcid{Qingyue Deng}{0009-0001-8471-2480},
  \authororcid{Jijiang Li}{0009-0003-2823-751X},
  \authororcid{Haibin Ling}{0000-0003-4094-8413}, and
  \authororcid{Bingyao Huang}{0000-0002-8647-5730}
}

\authorfooter{
  \item
    Haibo Li, Qingyue Deng, Jijiang Li, and Bingyao Huang are with Southwest University.
    E-mail: haiboli@email.swu.edu.cn, qingyue@email.swu.edu.cn, jijiangli@email.swu.edu.cn, bhuang@swu.edu.cn.
  \item Haibin Ling is with Westlake University.
  	E-mail: linghaibin@westlake.edu.cn
  \item
    Bingyao Huang is the corresponding author.
}

\teaser{
  \centering
  \includegraphics[width=\linewidth]{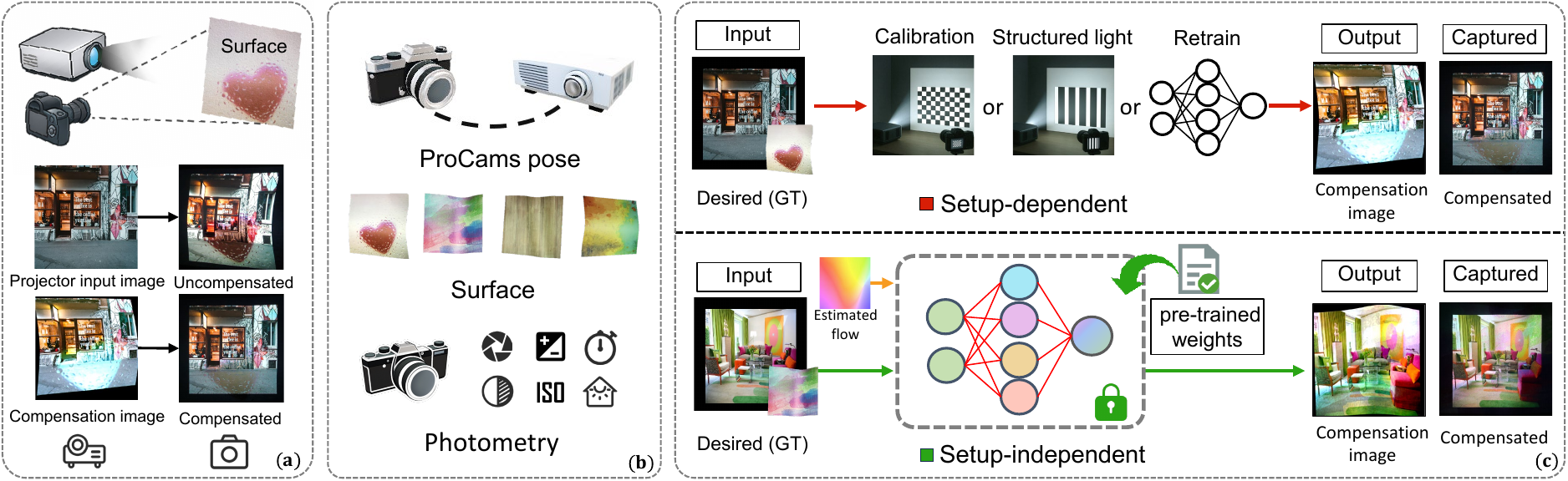}
  \caption{Setup-independent full projector compensation. (a) General projector compensation pipeline: if a projector directly projects an image onto a nonplanar and textured surface, the user may see a distorted (uncompensated) result, and the goal is to generate a compensation image such that when projected, it results in a visually pleasing (compensated) target appearance. (b) Traditional setup-dependent projector compensation methods cannot generalize to untrained new setups with various projector-camera poses, surface geometries, and photometric properties, such as exposure parameters (e.g., ISO, shutter speed) and ambient lighting conditions. (c) The top path depicts setup-dependent methods, which require per-setup calibration, reconstruction, or retraining for each new setup. In contrast, our setup-independent method (bottom path) is trained once on a diverse dataset; it only requires the estimated flow (detailed in~\cref{fig:SIComp_overall}) and pre-trained weights to generalize to unseen setups without further retraining or fine-tuning.}
  \label{fig:teaser}
}

\abstract{
Projector compensation seeks to correct geometric and photometric distortions that occur when images are projected onto nonplanar or textured surfaces. However, most existing methods are highly setup-dependent, requiring fine-tuning or retraining whenever the surface, lighting, or projector–camera pose changes. Progress has been limited by two key challenges: (1) the absence of large, diverse training datasets and (2) existing geometric correction models are typically constrained by specific spatial setups; without further retraining or fine-tuning, they often fail to generalize directly to novel geometric configurations. We introduce \textbf{SIComp}, the first \textbf{S}etup-\textbf{I}ndependent framework for full projector \textbf{Comp}ensation, capable of generalizing to unseen setups without fine-tuning or retraining. To enable this, we construct a large-scale real-world dataset spanning 277 distinct projector–camera setups. SIComp adopts a co-adaptive design that decouples geometry and photometry: A carefully tailored optical flow module performs online geometric correction, while a novel photometric network handles photometric compensation. To further enhance robustness under varying illumination, we integrate intensity-varying surface priors into the network design. Extensive experiments demonstrate that SIComp consistently produces high-quality compensation across diverse unseen setups, substantially outperforming existing methods in terms of generalization ability and establishing the first generalizable solution to projector compensation. The code and dataset are available on our project page: \url{https://hai-bo-li.github.io/SIComp/}.
}

\begin{document}
\bibliographystyle{abbrv-doi-hyperref}

\firstsection{Introduction}
\maketitle

Projectors are essential components in modern AR/VR systems, supporting a variety of applications ranging from interactive displays to projection mapping~\cite{bluteau2005visual,kanivets2020development,chien2022projection,edgcumbe2018follow, Yuta_Kageyama_2022_deblurring,ueda2020illuminated,Watanabe2024brightness_compensation,sato2024responsive,kageyama2024efficient,erel2024casper,takeuchi2024projection, Yuta2024BeamingDisplays,peng2025perceptually}. However, when projecting onto surfaces that are nonplanar and textured, the geometric and photometric distortions caused by surface geometry and reflectance significantly compromise visual quality. Projector compensation techniques aim to modify the projector input image to compensate for these distortions~\cite{siegl2015real,bermano2017makeup,grundhofer2018recent,cruzsurround,iwai2024projection,huang2019end,Kurth2018multiprojector_brightness_correction,yamamoto2022monocular}. As illustrated in~\cref{fig:teaser}(a), the camera-captured uncompensated image reveals significant geometric and photometric distortions, which are corrected in the compensated result.

A typical projector compensation system consists of a projector-camera pair that operates on a textured projection surface, as shown in~\cref{fig:teaser}(a). Throughout this paper, we use the term \textit{setup} to refer to a specific scene configuration, defined by the projection surface, the photometric conditions, and the poses of the projector–camera pair, as illustrated in \cref{fig:teaser}(b). Currently, numerous compensation methods~\cite{nayar2003projection,yoshida2003virtual,liu2011color,grundhofer2015robust,harville2006practical,huang2019end,wang2023compenhr,huang2021end} achieve effective results. However, their reliance on a fixed setup creates a significant bottleneck.
The core reason for setup dependency is that most previous methods rely on an auxiliary geometry correction process ~\cite{yoshida2003virtual,huang2019end,grundhofer2015robust} or require explicit calibration of the projector and camera~\cite{huang2021deprocams, Kurth2018calibration}. Although some methods~\cite{huang2019compennet++,huang2021end,wang2023compenhr} do not require calibration or structured light for geometric correction, their dependency on retraining or fine-tuning prevents the model from generalizing to new setups. The setup-dependent nature, which requires costly recalibration or retraining for any change in projector-camera pose, surface, or photometric condition, limits their scalability and practicality in dynamic environments.

Inspired by the strategy in CompenNeSt++~\cite{huang2021end}, which decouples the compensation problem into its geometric and photometric components, our work focuses on resolving their respective setup dependencies.

For geometric dependency, a seemingly straightforward approach is to leverage optical flow~\cite{lucas1981iterative} to directly estimate the projector-camera pixel mapping online. However, this is not feasible because optical flow assumes brightness constancy and small motion. This assumption does not hold for projector compensation, where a significant discrepancy exists between the projector input image and the uncompensated image captured by the camera, as shown in~\cref{fig:teaser}(a). To address this issue, we fine-tune a pre-trained learning-based optical flow model, FlowFormer~\cite{huang2022flowformer}, in conjunction with a pre-trained photometric compensation module, using our projector compensation dataset.
Consequently, our method eliminates the need for per-setup structured light or explicit calibration.

For photometric dependency, we find that typical CNN-based full projector compensation~\cite{huang2019compennet++,wang2023compenhr,huang2021end} approaches use a single camera-captured image of the projection surface to implicitly encode the properties of the surface and environmental lighting.
However, we argue that relying on a single surface image representation is often inadequate, particularly for complex or unseen setups with varying lighting conditions, as it may not sufficiently represent the photometric properties of the scene. Our subsequent analysis quantitatively reveals the limitations of this approach. Therefore, we use additional surface images under varying projector illumination levels. This enables the photometric module to more accurately model the dynamic range of the projector-camera system (ProCams) and the surface reflectance. Furthermore, we introduce an adaptive photometric module (IVPCNet) that integrates the Convolutional Block Attention Module (CBAM)~\cite{woo2018cbam} and the Swin Transformer~\cite{liu2021swin}. This design enables the architecture to effectively capture latent features of varying surfaces and photometric properties.

Another obstacle to achieving setup independence is the limited scale and diversity of the existing training dataset. A robustly generalizable model must be trained on data that represents a wide spectrum of real-world variations. To this end, we construct a large-scale dataset that specifically varies these critical factors, including ambient lighting, surface materials and reflectivities, and the geometric configurations of the ProCams.

Finally, we integrate the modules above and jointly optimize them using a co-adaptive training strategy on our large-scale and diverse training set. By working together, the modules constantly refine one another during training. Therefore, our SIComp can generalize to unseen projector-camera poses, photometric conditions, and projection surfaces without retraining or fine-tuning.
Our experimental results demonstrate that our setup-independent method, SIComp, shows highly competitive performance and even outperforms some setup-dependent approaches. Our key contributions are summarized below:
\begin{itemize}
  \item We propose SIComp, the first setup-independent full projector compensation framework capable of generalizing to unseen setups without retraining or fine-tuning.

  \item Our SIComp consists of a tailored optical flow module for online geometric correction under unseen setups and a photometric compensation module that leverages intensity-varying surfaces priors for more robust feature learning.

  \item We construct a large-scale and diverse dataset for full projector compensation comprising 277 unique setups. Experimental results demonstrate that our method significantly outperforms some setup-dependent baselines.
\end{itemize}

\section{Related Work}
\label{sec:relate}
\subsection{Setup-dependent projector compensation}
Projector compensation techniques are designed to compensate for the geometric and photometric distortions that arise from projecting onto non-ideal surfaces. Most research focuses on setup-dependent methods, which are intrinsically tied to a specific setup. Consequently, any change to the setup requires a recalibration or retraining. These methods are broadly classified into three categories: geometric correction, photometric compensation, and full compensation.

\subsubsection{Geometric correction}
Geometric correction aims to eliminate spatial distortions caused by nonplanar surfaces, ensuring that the projected image appears undistorted from the viewer’s perspective. Traditional strategies rely on structured light (SL) to establish camera-projector correspondences. Tardif~\etal~\cite{tardif2003multi} propose an SL-based mapping system for multi-projector setups that bypasses traditional calibration by directly recovering camera-projector mappings. Several studies have enhanced SL-based correction by improving projection accuracy, usability, and robustness~\cite{moreno2012simple, salvi2010state, wang2022high}. These methods achieve geometric correction by projecting and decoding a sequence of coded patterns. This reliance on pattern projection is often time-consuming and sensitive to environmental interference.

Marker-based approaches reduce calibration overhead by leveraging visual features. For instance, Audet~\etal~\cite{audet2009user} use printed markers, while Narita~\etal~\cite{narita2016dynamic} employ infrared (IR) markers for dynamic surfaces. To further enhance interaction, Kagami~\etal~\cite{kagami2019animated,kagami2020interactive} develop ``Animated/Interactive Stickies'' for low-latency mapping onto markerless planes. However, these methods are either restricted to planar surfaces or reliant on specialized hardware.

For complex deformations, Ibrahim~\etal~develop dynamic registration using Bézier patches~\cite{ibrahim2020dynamic,ibrahim2024multi} and self-calibrating extensions for jitter and occlusion~\cite{ibrahim2023projector, ibrahim2023self, ibrahim2024real}. These still rely on continuous ArUco marker visibility, which limits robustness on textured or rapidly stretching surfaces. To achieve markerless registration, Tehrani~\etal~\cite{tehrani2019automated} develop an automated system for arbitrary 3D shapes, but it requires a discrete re-estimation phase when projectors move, a process that typically takes several minutes.

Another geometric correction approach uses natural feature matching to bypass projected patterns. While the method by Takahashi~\etal~\cite{takahashi2010performance} enables correction from a single captured image, its reliance on 2D feature alignment limits its application to planar surfaces, as it cannot model 3D surface deformation.

\begin{table}[bt]
    \centering
    \caption{Comparison of full projector compensation methods. SI indicates Setup-Independent methods (\protect\usym{2713}).}
    \label{tab:method_comparison}
    \resizebox{\linewidth}{!}{
    \begin{tabular}{@{}l @{\hspace{1.5em}} c @{\hspace{0.8em}} c @{\hspace{0.8em}} c}
        \toprule
        \textbf{Methods} & \textbf{SI} & \textbf{Pre-calibration?} & \textbf{Setup time} \\
        \midrule
        Matrix-based~\cite{wetzstein2007radiometric, nayar2003projection} & \cellcolor{red!25}\usym{2717} & \cellcolor{red!25}Yes & \cellcolor{yellow!25}Medium \\
        CNN-based~\cite{wang2023compenhr,huang2021end} & \cellcolor{red!25}\usym{2717} & \cellcolor{green!25}No & \cellcolor{yellow!25}Medium \\
        Optical Flow-based~\cite{li2023physics} & \cellcolor{red!25}\usym{2717} & \cellcolor{green!25}No & \cellcolor{yellow!25}Medium \\
        Differentiable rendering-based~\cite{li2025dpcs,park2022projector} & \cellcolor{red!25}\usym{2717} & \cellcolor{red!25}Yes & \cellcolor{red!25}Long \\
        Radiance field-based~\cite{deng2025gs,erel2023nepmap} & \cellcolor{red!25}\usym{2717} & \cellcolor{red!25}Yes & \cellcolor{yellow!25}Medium \\
        SIComp (ours) & \cellcolor{green!25}\usym{2713} & \cellcolor{green!25}No & \cellcolor{green!25}Rapid \\
        \bottomrule
    \end{tabular}
    }
\end{table}

\subsubsection{Photometric compensation}
Photometric compensation aims to mitigate photometric distortions caused by textured projection surfaces and ambient lighting, typically assuming prior geometric correction. Early works~\cite{yoshida2003virtual, bimber2008visual, nayar2003projection, wang2005radiometric, fujii2005projector} often rely on linear models to simulate light interactions, where the contribution of each projector channel to each camera channel is approximated by a per-pixel $(3\times3)$ matrix. Fujii~\etal~\cite{fujii2005projector} demonstrate rapid frame-level adaptation to dynamic environmental changes, but their approach is limited in simultaneously handling complex reflectance and illumination variations, and its adaptability to new setups remains constrained. Yoshida~\etal~\cite{yoshida2003virtual} incorporate ambient light effects into photometric compensation; however, linear models struggle to represent the complex nonlinear photometric transformations encountered in real-world scenes.

Addressing shortcomings of the linear approaches above, nonlinear methods~\cite{liu2011color, grundhofer2013practical, grundhofer2015robust} based on lookup tables (LUTs) enable more accurate modeling of complex optical interactions. However, these methods require extensive calibration data and computational resources. For multi-projector setups, Pjanic~\etal~\cite{pjanic2018seamless} leverage the RLab color model to improve perceptual seamlessness, though it requires time-consuming LUT generation. Similarly, Tehrani~\etal~\cite{tehrani20233d} introduce 3D gamut morphing for color uniformity, but it lacks the capability to compensate for surface textures and is restricted to smooth surfaces without sharp normal variations.

More recently, deep learning approaches~\cite{huang2019end} show progress in learning complex color mappings without explicit calibration. Despite improved accuracy and reduced manual effort, these methods still require retraining for every new setup, making adaptation a time-consuming process.

\subsubsection{Full compensation}
Full compensation methods aim to correct geometric and photometric distortions simultaneously. Previous methodologies are characterized by specific technical assumptions and operational requirements. For instance, the light transport matrix method by Wetzstein~\etal~\cite{wetzstein2007radiometric} is designed for offline processing and involves significant computational time. For real-time applications, the system developed by Siegl~\etal~\cite{siegl2015real} operates based on the assumption that projection surfaces are Lambertian surfaces. Similarly, the approach by Shahpaski~\etal~\cite{shahpaski2017simultaneous} requires a pre-calibrated camera and a specific planar target, while its photometric compensation capabilities are constrained.

More recently, advanced data-driven and learning-based frameworks address geometric and photometric compensation problems jointly. These frameworks leverage techniques such as differentiable rendering, neural rendering, and deep neural networks.

Deep convolutional neural networks-based methods~\cite{huang2021end, wang2023compenhr} learn corrections directly from data, avoiding the need for explicit calibration. However, their applicability is constrained to static setups. Even with rapid fine-tuning, the process takes several minutes, which impedes real-time deployment.
Recently, some methods~\cite{li2023physics, wang2024vicomp} leverage optical flow to directly estimate the dense pixel mapping between the projector and the camera images, without baking the geometric correction in the network parameters. However, they still require a multi-minute adaptation phase for new setups.

Differentiable rendering frameworks are proposed to decouple light, material, and shape from the compensation process. DeProCams~\cite{huang2021deprocams} introduces a neural rendering framework that simultaneously performs ProCams relighting, compensation, and shape reconstruction. Park~\etal~\cite{park2022projector} leverages differentiable rendering for iterative refinement-based compensation. DPCS~\cite{li2025dpcs} uses physically-based path tracing to model the light transport between the projector and the camera, and leverages differentiable rendering to compensate for projector images. While powerful for modeling complex light transport and scene details, these methods require high computational costs and typically rely on accurate initial calibration.

Radiance field-based frameworks are also applied to projector compensation. Erel~\etal~\cite{erel2023nepmap} propose the first neural reflectance fields-based ProCams model, named NepMap. Although powerful, it requires massive GPU memory and training time. Deng~\etal~\cite{deng2025gs} introduce the first Gaussian splatting-based ProCams model. By representing the scene with 2D Gaussians~\cite{huang20242d}, GS-ProCams enables efficient, view-agnostic projection mapping. However, both methods still require a per-scene optimization from multi-view captures.

In summary, prior works (\cref{tab:method_comparison}) suffer from a rigid dependency on system configuration. Since any change to the hardware requires an adaptation, recalibration, or retraining cycle, we focus on developing a projector compensation method that is truly setup-independent.

\subsection{Techniques used in our work}
To the best of our knowledge, our work, SIComp, is the first setup-independent full projector compensation method. We achieve this by designing a novel framework that integrates a pre-trained optical flow module for online geometry correction, and a pre-trained photometric compensation network with intensity-varying surface priors and attention mechanisms.

\paragraph{Optical flow.} Recent advances in optical flow methods, such as RAFT~\cite{teed2020raft} and FlowFormer~\cite{huang2022flowformer}, enable highly accurate and robust pixel-level correspondence estimation without predefined patterns. This capability presents a promising alternative to structured light for establishing the geometric mapping between the projector and camera, forming a solid foundation for calibration-free geometric correction.

\paragraph{Attention mechanisms.} Modules like the Swin block~\cite{liu2021swin} and CBAM~\cite{woo2018cbam} have become a cornerstone of deep learning due to their ability to dynamically weigh the importance of different spatial and channel-wise features. This adaptivity empowers a single projector compensation model to better learn diverse surface textures and photometric conditions without being explicitly trained for each setup.

\begin{figure*}[ht]
  \centering
  \includegraphics[width=1\linewidth]{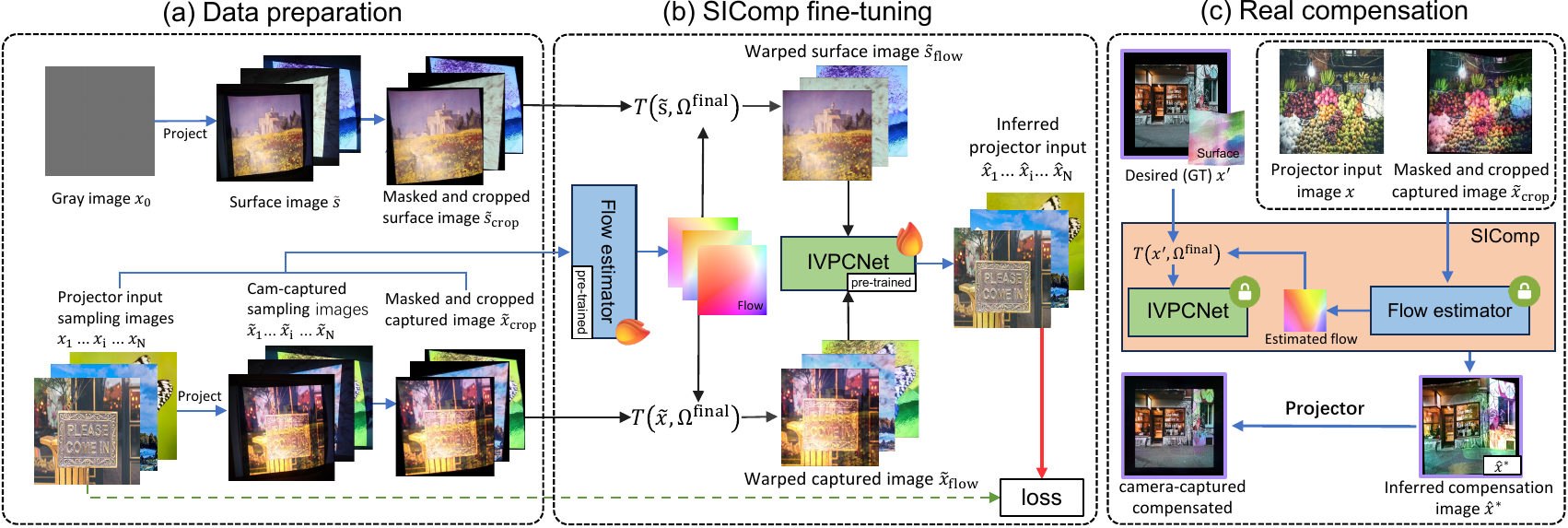}
  \caption{SIComp pipeline. (a) Data preparation phase, including the acquisition of surface images and various captured projection images collected under diverse setups, followed by masking and cropping. (b) SIComp fine-tuning pipeline. It comprises a flow estimator module for geometric correction, which is pre-trained on the Sintel~\cite{Butler:ECCV:2012} dataset. A pre-trained IVPCNet performs photometric compensation, inferring the projector input image based on optical flow warped images, whose pre-training process is detailed in~\cref{fig:IVPCNet}. (c) Real compensation application, where the trained SIComp takes a desired image $x'$ and the optical flow (pre-estimated from an initial reference projection) as input to infer the compensation image, which is then physically projected and captured by a camera to approximate the desired image $x'$. The symbol $T$ denotes the warping operation applied using the estimated optical flow.}
\label{fig:SIComp_overall}
\end{figure*}

\begin{figure*}[htbp]
  \centering
  \includegraphics[width=\textwidth]{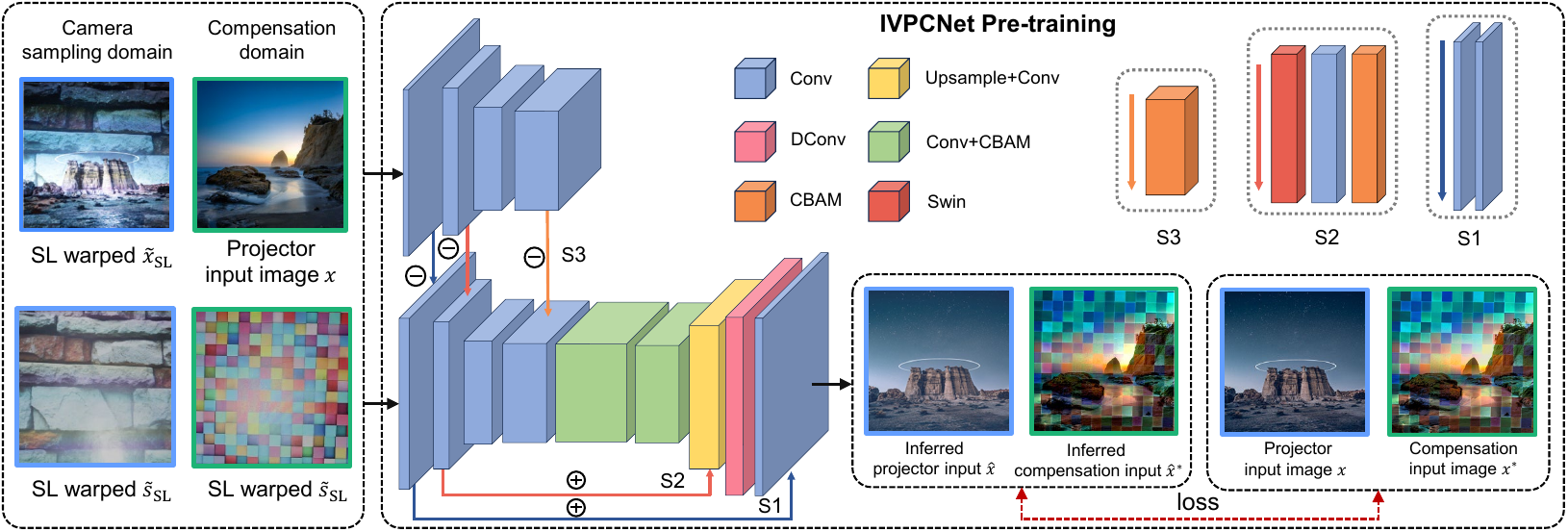}
\caption{Pre-training pipeline for IVPCNet.
IVPCNet, a siamese U-Net architecture, is pre-trained using two different domains.
The first is the \textbf{camera sampling domain} (solid blue border). It takes a camera-captured image warped by structured light and the corresponding surface image as input, predicting an output $\hat{x}$.
The second is the \textbf{compensation domain} (solid green border), where the network learns to predict a compensation image. It uses the projector input image $x$ and the same warped surface image to predict $\hat{x}^{*}$, with the loss computed against a target compensation image $x^{*}$. This target image $x^{*}$ is derived for the specific hardware setup using Nayar's iterative refinement method~\cite{nayar2003projection}.
This pre-training process yields the IVPCNet model for subsequent fine-tuning within the SIComp framework (see~\cref{fig:SIComp_overall}). The blue, pink, and orange lines denote the skip connections at stages s1, s2, and s3, respectively.}  \label{fig:IVPCNet}
\end{figure*}

\section{Method}
\label{sec:Method}
\subsection{Problem formulation}
\label{sec:problem_formulation}
The primary goal of projector compensation is to project light in such a way that the resulting image, as perceived by an observer in the real world, appears geometrically correct and photometrically faithful, regardless of the surface's shape or texture.

To formalize this challenge, we model the physical process within a projector-camera system, where a camera acts as an effective proxy for a human observer. Consider a setup where the intended projector input image, denoted by $x$, is sent to the projector and is directly projected onto the surface without any compensation. The camera typically captures a distorted version of this input, which we label as the uncompensated image $\tilde{x}$ (see~\cref{fig:teaser}(a)). This distortion stems from two primary sources in the image formation process: a \textbf{photometric transformation $R$}, which accounts for surface reflectance and ambient lighting $E$ interactions, and a \textbf{geometric mapping $G$}, which models the spatial warping caused by the surface's shape and the projector-camera alignment. Let $S$ denote the combined properties of the surface (geometry and reflectance). The entire process that maps the ideal input $x$ to the captured distorted image $\tilde{x}$ can be expressed as:

\begin{equation}
\tilde{x} = G \left( R \left( x; E, S \right) \right).
\label{eq:forward_process}
\end{equation}

Consequently, the objective of projector compensation is to find an adjusted compensation image, $\hat{x}^{*}$, for projection. This image is designed to counteract the distortions that occur when it is projected onto a surface and subsequently captured by a camera. We assume there exists a perfect compensation image, $x^*$, when subjected to the same degradation projection-capture pipeline, completely neutralizes all distortions, leading to the following theoretical relationship, reproducing the original projector input image $x$. This ideal compensation process is defined by the inverse relationship:

\begin{equation}
x = G \left( R \left( x^*; E, S \right) \right).
\label{eq:compensation_process}
\end{equation}

However, directly measuring the complex interplay of surface properties $S$ and ambient illumination $E$ is often impractical. Following~\cite{huang2021end}, we instead utilize a camera-captured image $\tilde{s}$ (surface image in \cref{fig:SIComp_overall}(a)) to implicitly encapsulate the combined effects of $E$ and $S$. This surface image $\tilde{s}$ is obtained by projecting a uniform gray image ($x_{0}$ in~\cref{fig:SIComp_overall}(a)) onto the surface under both ambient and projector illumination:

\begin{equation}
\tilde{s} = G \left( R \left( x_0; E, S \right) \right),
\label{eq:surface_image}
\end{equation}
where $x_0$ is a plain gray image providing a reference illumination.

To derive the compensation image $x^*$, we aim to approximate the inverse of the composite function in \cref{eq:compensation_process}. However, due to the inherent nonlinear characteristics of both the projector and the camera’s color response functions, obtaining an exact inverse of $R$ is generally infeasible. Therefore, we introduce a pseudo-inverse operation $R^{\dagger}$, which represents the approximation of the inverse photometric transformation. Then the ideal compensation image $x^*$ is expressed as:

\begin{equation}
x^* = R^{\dagger} \left( G^{-1}(x), \ G^{-1}(\tilde{s}) \right),
\label{eq:get_cmp_image}
\end{equation}

\noindent where $R^{\dagger}$ denotes the pseudo-inverse of $R$, and $G^{-1}$ represents the inverse geometric mapping that aligns images from the camera view back to the projector’s coordinate space.

\subsection{Definition of unseen setup}
\label{sec:def_unseen_setup}
In this work, a setup denotes a specific configuration characterized by three core factors: (1) the projection surface, encompassing its geometry and reflectance properties; (2) the spatial relationship (relative pose) between the projector and the camera; and (3) the photometric conditions during capture, which include ambient illumination and camera imaging parameters (e.g., exposure, white balance). A key challenge in projector compensation is that these factors are coupled in the captured image. Our method is setup-independent as it is trained to disentangle these factors and can generalize to unseen setups, i.e., novel, previously unseen combinations of the above elements, without requiring any per-setup retraining or fine-tuning.

\subsection{SIComp: setup-independent compensation}
\label{sec:learning_framework}
Distinct from prior methods, our SIComp is specifically designed to jointly correct for both geometric and photometric distortions in a setup-independent manner, allowing it to adapt to new setups without retraining. We implement~\cref{eq:get_cmp_image} via two modules, each dedicated to a specific part of the decomposition.

For geometric correction, inspired by previous works~\cite{li2023physics,wang2024vicomp}, we leverage an online optical flow estimator. Its ability to predict dense pixel correspondences without relying on explicit projector-camera poses is ideal for our setup-independent goal. This module's primary role is to compute the inverse geometric mapping $G^{-1}$ by predicting a flow field (or displacement field) $\Omega$. This design directly addresses a key limitation of previous approaches like WarpingNet~\cite{huang2021end} and GANet~\cite{wang2023compenhr}, which fail to generalize to new setups.

For photometric compensation, we utilize the geometrically corrected images from the previous stage as input. This module's task is to learn and approximate the pseudo-inverse operation $R^{\dagger}$, compensating for the complex photometric distortions caused by textured surfaces and photometric conditions.

By integrating these two components, SIComp directly models the geometric and photometric inverse operations. This process can be described as:

\begin{equation}
x^* = R^\dagger\left( T \left( x, \Omega \right), T \left( \tilde{s}, \Omega \right) \right),
\label{eq:compensation-with-optical}
\end{equation}
where $T$ represents the operation of warping an image using the flow field $\Omega$.

\subsubsection{Surrogate training for compensation}
\label{sec:preliminaries}
A challenge in learning-based compensation methods is the lack of ground truth for the ideal compensation image $x^{*}$. To address this, CompenNet~\cite{huang2019end} proposes a surrogate training strategy: instead of directly learning the true mapping ($x \rightarrow x^{*}$), the network learns to recover the original input image $x$ from the uncompensated image $\tilde{x}$, i.e., ($\tilde{x} \rightarrow x$). This approach allows for easily collected training pairs $(\tilde{x}, x)$, enabling effective supervision and ensuring that the network can perform the forward compensation task during inference.

\label{sec:optimization_objective}
Following the surrogate training principle above, we train SIComp to perform the inverse task of recovering the original projector input image. We define the overall SIComp network as $O_{g, r}$, parameterized by its geometric ($g$) and photometric ($r$) components. The network $O_{g,r}$ takes an uncompensated image $\tilde{x}$ and surface image $\tilde{s}$ as input to predict the original projector image, denoted as $\hat{x}$:
\begin{equation}
    \hat{x} = O_{g, r}\left( \tilde{x}, \tilde{s} \right).
    \label{eq:network-inverse}
\end{equation}

Accordingly, the trainable parameters $g$ and $r$ are optimized by minimizing a loss function $\mathcal{L}$ that measures the discrepancy between the network's prediction $\hat{x}$ and the ground truth projector input image $x$. The overall training objective for a mini-batch of size $B$ is formally expressed as:

\begin{equation}
    \min_{g, r} \frac{1}{B} \sum_{i=1}^{B} \mathcal{L}\left(O_{g, r}\left(\tilde{x}^{(i)}, \tilde{s}^{(i)}\right),\, x^{(i)}\right),
    \label{eq:training-loss}
\end{equation}
where $\mathcal{L}$ represents a composite reconstruction loss, which will be detailed in~\cref{loss_function}.

\subsubsection{Detailed training strategy}
\label{train_strategy}
While the surrogate training objective is well-defined, optimizing it directly on raw camera-captured, uncompensated images is challenging. These inputs suffer from significant background clutter and photometric inconsistencies, which can destabilize training, particularly for the optical flow subnet. To address this, we introduce a three-part strategy involving input pre-processing, introducing an intensity-varying surface prior, and two-stage optimization.

\begin{figure}[tb]
  \centering
  \includegraphics[width=1\linewidth]{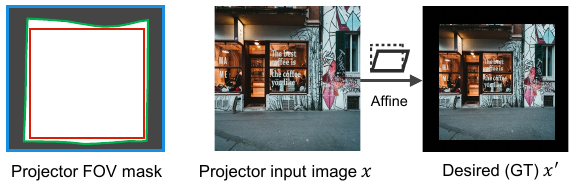}
  \caption{Projector and camera field of view (FOV). In the projector FOV mask, the white region represents the projector's FOV. The camera's FOV encompasses the broader scene, represented by the blue region, and the optimal viewing area is highlighted by the red boundary. Due to the projector's FOV and potential geometric distortions of the surface, an affine transformation is applied to the projector input image, resulting in the ideal visualization after accounting for the projector's FOV limitations and the surface geometric distortions, i.e., desired ground truth (GT).
}
  \label{fig:projector_fov}
\end{figure}

\paragraph{Input masking and cropping.}
\label{sec:mask_crop}
The first step in our strategy is to eliminate irrelevant background regions from the captured images, which act as a major source of noise for geometric alignment. We employ a two-step process to isolate the valid projector field of view (FOV). First, a binary mask is applied to the uncompensated image $\tilde{x}$ and the surface image $\tilde{s}$ to remove all areas outside the projector's FOV as illustrated in~\cref{fig:projector_fov}. Then,
we tightly crop the masked images to the bounding box of the projector's FOV. This yields clean, content-focused inputs, $\tilde{x}_{\text{crop}}$ and $\tilde{s}_{\text{crop}}$, which significantly improve the accuracy of flow prediction and accelerate network convergence. The result of this process is visualized in~\cref{fig:SIComp_overall}(a).

\paragraph{Optimal visualization area.}
\label{sec:Optimal-Visualization-Area}
During training, the network is supervised using captured uncompensated images $\tilde{x}$, which inherently account for the surface's geometric distortion and are not limited by the projector's field of view. In contrast, at real compensation inference time, directly inputting the projector input image $x$ to obtain the compensation image $\hat{x}^*$ will result in a suboptimal compensation result. To prevent clipping or content loss during projection, we apply an affine transformation to the projector input $x$, producing a transformed version $x'$ (Desired (GT)) that fits within the largest inscribed rectangle of the projector’s field of view (~\cref{fig:projector_fov}). This ensures that the predicted compensation image $\hat{x}^*$ is fully visible when projected. During inference, we use $x'$ as the input to generate the compensation image $\hat{x}^{*}$.

\paragraph{Intensity-varying surface prior.}
\label{sec:multi_surface_augmentation}
To improve the generalization ability of SIComp to varying surface illumination levels, we introduce an intensity-varying surface prior. Unlike previous methods~\cite{huang2021end,wang2023compenhr} that provide the network with a single surface image ($\tilde{s}_{\text{crop}}$), we augment the input with a sequence of surface captures that profile the surface's response to different projector illumination intensities.

\begin{figure}[tb]
  \centering
  \includegraphics[width=1\linewidth]{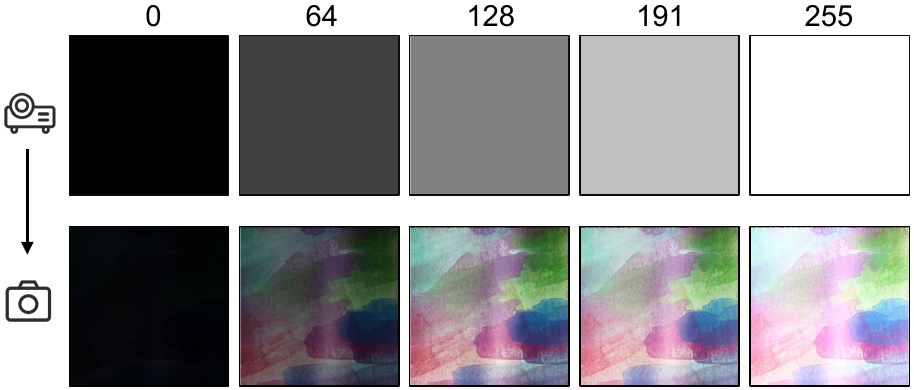}
  \caption{Intensity-varying surface priors. (Top) Uniform gray images of varying intensities (0, 64, 128, 191, and 255) are projected onto a surface. (Bottom) The corresponding camera captures reveal the surface's distinct reflectance properties under various projection intensities. This set of captured images provides intensity-varying surface priors, enabling SIComp to more effectively learn complex surface properties.}
  \label{fig:multi_surface}
\end{figure}

We project uniform grayscale images at a set of five discrete intensity levels: $\{0, 64, 128, 191, 255\}$, spanning the projector's dynamic range (~\cref{fig:multi_surface}). This yields five fixed surface images per setup. We experiment with three specific input configurations defined by a hyperparameter $K$, each using a fixed set of intensities throughout training and evaluation: for $K=1$ only intensity $64$; for $K=3$ intensities $\{0, 128, 255\}$; and for $K=5$ all five intensities. These intensity-varying surface priors enable the network to learn a more robust compensation model.

\paragraph{Pre-training and fine-tuning.}
\label{sec:staged_optimization}
Training SIComp from scratch is challenging due to the complexity of jointly learning geometric and photometric transformations. We address this with a two-stage optimization strategy.
In the first stage, the geometric module is pre-trained on the Sintel~\cite{Butler:ECCV:2012} dataset, while the photometric module is pre-trained on geometrically corrected image pairs from real-world setups to model photometric inconsistencies. These pre-trained subnets provide a robust foundation for the second stage, where both modules are fine-tuned jointly on real-world geometrically-uncorrected data. This fine-tuning process enables both modules to co-adapt to the final full projector compensation task. During this phase, SIComp takes the pre-processed inputs defined in the ~\cref{train_strategy} to predict the original projector image. The final input is given by:

\begin{equation}
    \hat{x} = O_{g, r}\left(\tilde{x}_{\text{crop}}, \mathcal{P}_K \right), \quad K \in \{1, 3, 5\}, \label{eq:network-training-input}
\end{equation}
where $K \in \{1, 3, 5\}$ specifies the number of intensity-varying surface priors included in the input set $\mathcal{P}_K$.

During real compensation, the desired GT $x'$ and the pre-estimated optical flow are fed into the trained SIComp to obtain the predicted compensation image $\hat{x}^*$. Then $\hat{x}^*$ is projected onto the surface. The camera captured result of this projection effectively compensated both geometric and photometric distortions of the scene, resulting in a visually corrected image that closely matches the desired GT $x'$. The full process is illustrated in \cref{fig:SIComp_overall}(c).

For each new setup, a one-time preparation is performed: a reference image is captured to pre-estimate the optical flow, followed by the acquisition of the five intensity-varying surface priors. This entire process takes approximately 2-3 seconds. During inference, the model utilizes the estimated flow field and the subset $\mathcal{P}_K$ (where $K \in \{1,3,5\}$) corresponding to its trained configuration. Since these components are stored and reused for all subsequent compensations on the same setup, this one-time data capture overhead is excluded from the reported per-image inference times.

\subsection{Network architecture}
The SIComp framework consists of two main modules: an optical flow estimator module based on FlowFormer~\cite{huang2022flowformer}, and a photometric compensation module based on a modified siamese U-Net, which is named Intensity-Varying Photometric Compensation Network (IVPCNet).

\subsubsection{Geometric correction via FlowFormer}
FlowFormer~\cite{huang2022flowformer} estimates an optical flow field $\Omega$ between the cropped camera image $\tilde{x}_{\text{crop}}$ and the projector input image $x$. This flow field represents the pixel-wise displacements from the camera viewpoint to that of the projector.

The geometric correction process within FlowFormer is performed in two stages:
(1) A Transformer-based cost volume encoder embeds a 4D cost volume into a latent space, forming a global cost memory.
(2) A decoder retrieves information from this memory using cost queries and iteratively predicts residual flows $\Delta\Omega(t)$ via recursive attention and a ConvGRU module. Starting from an initial estimate $\Omega(0)=0$, the flow estimate is updated recursively as $\Omega(t+1) \leftarrow \Omega(t) + \Delta\Omega(t)$. After $N$ iterations, the final accumulated flow field is $\Omega^{\text{final}} = \sum_{t=1}^{N} \Delta\Omega \left( t \right)$.
We then use this final flow field, $\Omega^{\text{final}}$, to warp the cropped camera image $\tilde{x}_{\text{crop}}$, producing the geometrically corrected image $\tilde{x}_{\text{flow}}$. The same flow field is then applied to the cropped surface images $\tilde{s}_{\text{crop}}$ to yield $\tilde{s}_{\text{flow}}$, as illustrated in \cref{fig:SIComp_overall}(b).

\subsubsection{Photometric compensation via IVPCNet}
The IVPCNet architecture is shown in \cref{fig:IVPCNet}, and it corrects photometric distortions in the flow-warped image $\tilde{x}_{\text{flow}}$ to produce the compensated output $\hat{x}$. To do so, it takes the $\tilde{x}_{\text{flow}}$ and the corresponding aligned surfaces $\tilde{s}_{\text{flow}}$ as input. The design incorporates three key components to achieve generalized photometric compensation:
(1) \textbf{Intensity-varying surface priors input design}, which enables the network to learn different surface photometric variations and improve generalization across diverse surfaces.
(2) \textbf{Swin~\cite{liu2021swin} Transformer Attention}, applied in skip connections with shifted windows. This mechanism captures long-range dependencies through Query-Key-Value (QKV) similarity, enhancing the network's ability to model complex photometric conditions.
(3) \textbf{Dynamic Feature Gating} via CBAM~\cite{woo2018cbam}, applied after each encoder and decoder block. This module adaptively combines channel and spatial attention to emphasize critical regions for photometric correction.

\begin{figure*}[htb]
  \centering
   \includegraphics[width=.9\linewidth]{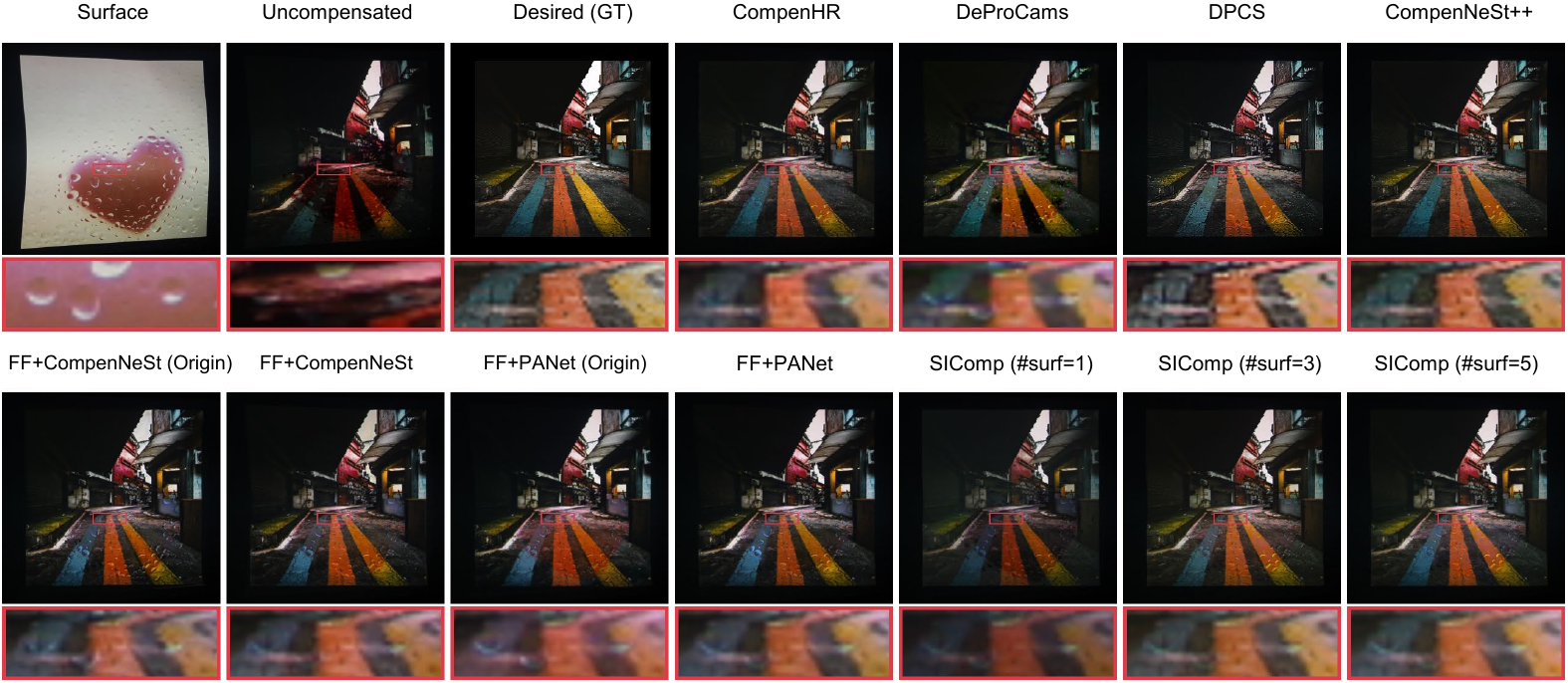}
    \includegraphics[width=.9\linewidth]{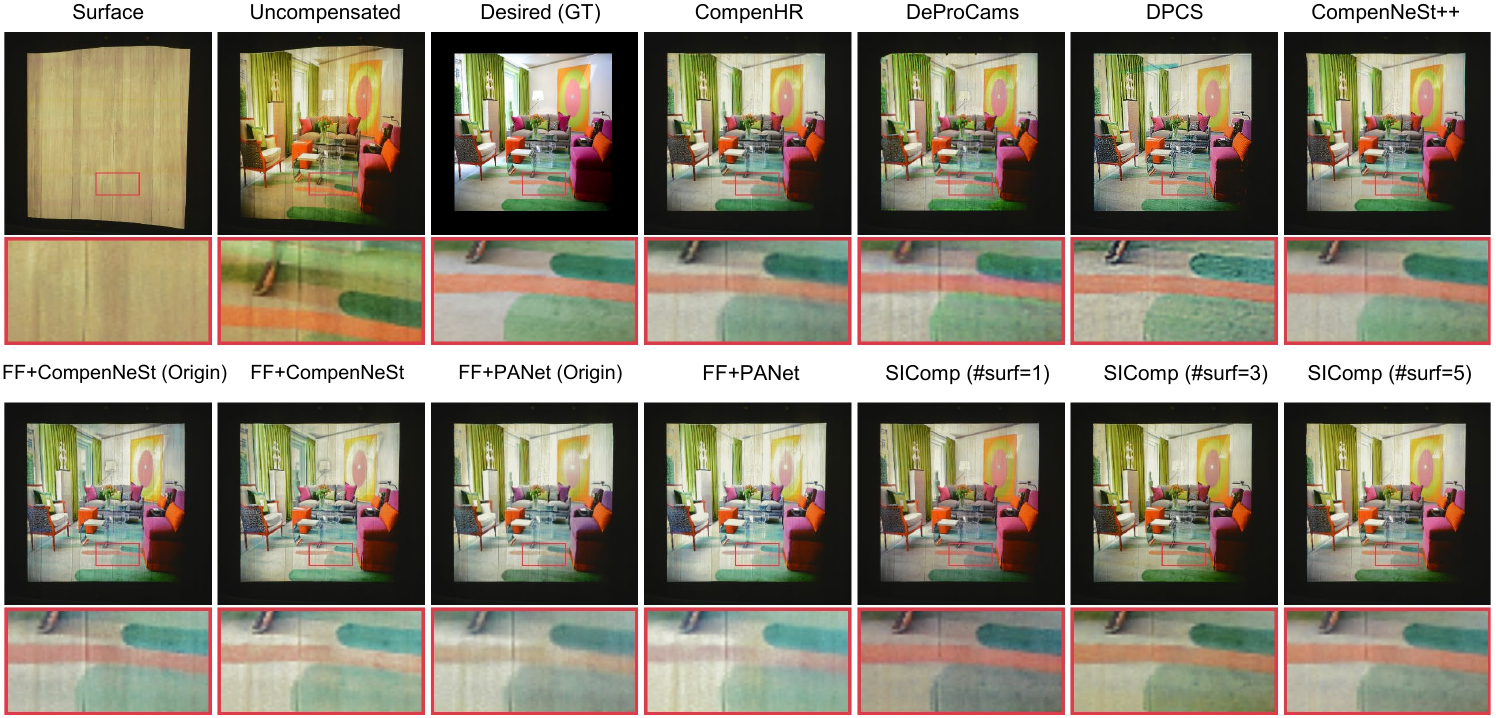}
  \caption{Qualitative results of real compensation experiments. \textbf{Rows 1–2 (Set A)}: same ProCams devices, unseen setups. \textbf{Rows 3–4 (Set B)}: novel ProCams devices, unseen setups. For each pair of rows: The top row displays the surface, uncompensated image, desired ground truth, and setup-dependent results (CompenHR, DeProCams, DPCS, CompenNeSt++); the bottom row shows setup-independent results (FF+CompenNeSt, FF+PANet, and SIComp with 1, 3, and 5 surfaces). Red boxes indicate magnified insets for comparison. See more results in the Supplementary Material.}
  \label{fig:real_cmp_results}
\end{figure*}

\section{Experiment}
To evaluate the generalization ability of SIComp, we conduct two types of experiments under unseen setups: real compensation and surrogate compensation.
Real compensation involves physically projecting the inferred compensation image onto a surface and comparing the camera-captured result with the desired (GT). This protocol facilitates a more realistic and comprehensive assessment of the model’s effectiveness under practical, unconstrained conditions.

We benchmark SIComp against several projector compensation methods, broadly categorized into two main groups based on their adaptability to unseen setups: setup-dependent and setup-independent.

\textbf{Setup-dependent methods}: This group includes DPCS~\cite{li2025dpcs}, DeProCams~\cite{huang2021deprocams}, CompenNeSt++~\cite{huang2021end}, and CompenHR~\cite{wang2023compenhr}. These models require retraining or extensive optimization for each new surface, photometric condition, or projector-camera pose.

\textbf{Setup-Independent (SI) methods}: This group includes FF+PANet, FF+CompenNeSt, and our proposed SIComp variants. We train these models only once on a diverse dataset, allowing them to work directly on unseen setups without any retraining or fine-tuning.
To create strong setup-independent baselines, we combine a shared FlowFormer (FF) with the photometric networks from CompenNeSt++~\cite{huang2021end} and CompenHR~\cite{wang2023compenhr}, forming FF+CompenNeSt and FF+PANet, respectively. To ensure a fair comparison, we evaluate these baselines under two settings: (1) their original, published configurations (denoted as (Origin)), and (2) variants that we retrain using our unified training setup (see~\cref{train_config}) to isolate the impact of the training strategy. Specifically, the (Origin) versions omit photometric pre-training and follow the original fine-tuning hyperparameters from their respective publications (with the batch size and learning rate schedule linearly scaled for FF+CompenNeSt to reduce GPU memory usage).

Our ablation study includes three SIComp variants, differing only in the number of input surface priors (1, 3, and 5), to evaluate the impact of the number of surface priors on model generalization.

\subsection{Datasets and implementation details}

\subsubsection{Datasets}
\label{dataset}
To pre-train IVPCNet and fine-tune SIComp, we construct a large-scale real-world dataset consisting of 277 distinct projector-camera setups. We collect 223 of these setups through a structured data acquisition process involving real projections and camera captures. The remaining 54 setups come from existing public datasets: nine from CompenNeSt++~\cite{huang2021end}, 21 from CompenHR~\cite{wang2023compenhr}, and 24 from CompenNet~\cite{huang2019end}.

Each setup contains 500 training images and 200 validation images, totaling approximately 138,500 training samples and 55,400 validation samples. All training and validation image resolutions are $256 \times 256$ pixels. These setups include a wide range of variations in surface texture (e.g., smooth, rough, patterned), reflectance properties (e.g., matte, glossy), projection angles and distances, and environmental lighting conditions (e.g., varying ambient light, exposure time, and ISO settings).

Each data sample includes three components: (1) the camera-captured images; (2) multiple surface images captured under pure-gray projections at varying intensity levels, serving as surface priors; and (3) the projector input images. These components together allow the network to learn both geometric and photometric compensation across diverse real-world setups. The comprehensive dataset provides sufficient diversity and scale to support robust training of setup-independent projector compensation models.

To evaluate the generalization ability of SIComp, we create two distinct evaluation sets with a resolution of $600 \times 600$ pixels, significantly higher than the training resolution $(256 \times 256)$, thereby assessing the models' ability to generalize across different image scales. Specifically, we establish two datasets for real compensation evaluation:

\noindent\textbf{(1) Set A: same ProCams devices and unseen setups}: It comprises 8 unseen setups (4 surfaces $\times$ 2 poses), captured using a Canon 600D camera and an EPSON CB-965 projector, with 5 test images per setup.

\noindent\textbf{(2) Set B: novel ProCams devices and unseen setups}: This set evaluates generalization  to novel hardware and setups (\cref{tab:real_cmp_results}). It consists of 5 unseen setups across 3 diverse physical surfaces and 5 unique projector-camera poses, captured using a Nikon D3200 camera and a TOSHIBA TDP-T100C projector, with 5 test images per setup. Note that we do not use this ProCams device for any training data.

\subsubsection{Training configurations}
\label{train_config}
Experiments are conducted on a workstation featuring two NVIDIA RTX 3090s, one RTX 4080, and one A100 GPU. All models are trained for 12,000 iterations using the Adam~\cite{kingma2015adam} optimizer with a batch size of 6. FlowFormer and IVPCNet use initial learning rates of $3.5 \times 10^{-5}$ and $1.0 \times 10^{-4}$ respectively, both with a weight decay of $1.0 \times 10^{-5}$. A step decay strategy is applied every 5,000 iterations, reducing the learning rates by factors of 0.90 for FlowFormer and 0.3 for IVPCNet.

\subsubsection{Evaluation metrics and loss function}
\label{loss_function}
To quantitatively assess the performance of each projector compensation method, we employ several widely recognized metrics: Peak Signal-to-Noise Ratio (PSNR), Root Mean Square Error (RMSE), Structural Similarity Index Measure (SSIM)~\cite{wang2004image}, Color Difference ($\Delta$E, CIEDE2000)~\cite{sharma2005ciede2000}, Learned Perceptual Image Patch Similarity (LPIPS)~\cite{zhang2018unreasonable}, and Fréchet Inception Distance (FID)~\cite{heusel2017gans}.

We train SIComp using an $\ell_1 + \text{SSIM}$ loss combination, which yields the strongest performance on the primary pixel-based metrics (PSNR and SSIM). The comparison of different loss functions is shown in the Supplementary Material.

\begin{table*}[t]
\centering
\caption{Quantitative comparison of \textbf{real} projector compensation methods on \textbf{Set A} (same ProCams devices, unseen setups, averaged over 8 setups) and \textbf{Set B} (novel ProCams devices, unseen setups, averaged over 5 setups). SI indicates Setup-Independent methods (\protect\usym{2713}). The best results within each category are \textbf{bolded}.}
\label{tab:real_cmp_results}
\resizebox{.9\linewidth}{!}{
\setlength{\tabcolsep}{2.2pt}
\begin{tabular}{@{}lc @{\hspace{6pt}} cccrcr @{\hspace{6pt}} rccrcr@{}}
\toprule
& & \multicolumn{6}{c}{\textbf{Set A}: same ProCams devices, unseen setups} & \multicolumn{6}{c}{\textbf{Set B}: novel ProCams devices, unseen setups} \\
\cmidrule(lr){3-8} \cmidrule(lr){9-14}
\textbf{Method} & \textbf{SI} & \textbf{PSNR}$\uparrow$ & \textbf{RMSE}$\downarrow$ & \textbf{SSIM}$\uparrow$ & $\Delta$\textbf{E}$\downarrow$ & \textbf{LPIPS}$\downarrow$ & \textbf{FID}$\downarrow$ & \textbf{PSNR}$\uparrow$ & \textbf{RMSE}$\downarrow$ & \textbf{SSIM}$\uparrow$ & $\Delta$\textbf{E}$\downarrow$ & \textbf{LPIPS}$\downarrow$ & \textbf{FID}$\downarrow$ \\
\midrule
CompenNeSt++~\cite{huang2021end} & \usym{2717} & \textbf{24.64} & \textbf{0.063} & \textbf{0.883} & \textbf{4.48} & \textbf{0.110} & \textbf{72.77} & \textbf{23.75} & \textbf{0.067} & \textbf{0.802} & \textbf{4.98} & \textbf{0.156} & \textbf{96.11} \\
CompenHR~\cite{wang2023compenhr} & \usym{2717} & 23.69 & 0.069 & 0.851 & 5.33 & 0.153 & 91.52 & 23.74 & \textbf{0.067} & 0.797 & 5.55 & 0.178 & 99.56 \\
DeProCams~\cite{huang2021deprocams} & \usym{2717} & 22.32 & 0.080 & 0.807 & 6.25 & 0.202 & 119.39 & 21.96 & 0.082 & 0.740 & 6.65 & 0.226 & 128.29 \\
DPCS~\cite{li2025dpcs} & \usym{2717} & 22.48 & 0.078 & 0.819 & 6.06 & 0.165 & 92.70 & 20.53 & 0.097 & 0.715 & 8.14 & 0.250 & 140.55 \\
\midrule
FF+CompenNeSt & \usym{2713} & 20.76 & 0.097 & 0.826 & 6.95 & 0.156 & 119.63 & 21.48 & 0.087 & 0.779 & 7.26 & 0.165 & 123.00 \\
FF+CompenNeSt (Origin) & \usym{2713} & 19.17 & 0.115 & 0.704 & 7.50 & 0.182 & 120.34 & 19.43 & 0.110 & 0.641 & 8.20 & 0.208 & 127.61 \\
FF+PANet & \usym{2713} & 21.78 & 0.085 & 0.823 & 6.70 & 0.170 & 117.32 & 21.20 & 0.090 & 0.759 & 7.95 & 0.188 & 131.09 \\
FF+PANet (Origin) & \usym{2713} & 21.13 & 0.091 & 0.799 & 8.21 & 0.206 & 135.04 & 20.95 & 0.091 & 0.735 & 8.25 & 0.217 & 134.52 \\
SIComp (\#surf=1) & \usym{2713} & 18.23 & 0.134 & 0.752 & 10.02 & 0.191 & 140.45 & 22.07 & 0.083 & 0.788 & 7.53 & 0.153 & 98.57 \\
SIComp (\#surf=3) & \usym{2713} & 22.79 & 0.076 & 0.840 & 6.24 & \textbf{0.144} & 95.07 & 23.11 & 0.071 & 0.793 & 6.06 & 0.157 & 100.34 \\
SIComp (\#surf=5) & \usym{2713} & \textbf{23.51} & \textbf{0.070} & \textbf{0.846} & \textbf{5.76} & 0.146 & \textbf{90.95} & \textbf{23.46} & \textbf{0.069} & \textbf{0.802} & \textbf{5.87} & \textbf{0.148} & \textbf{89.60} \\
\bottomrule
\end{tabular}
}
\end{table*}

\begin{table}[ht]
\centering
\caption{Quantitative comparison of \textbf{surrogate} projector compensation methods on \textbf{Set A} (same ProCams devices, unseen setups, averaged over 8 setups). SI indicates Setup-Independent methods (\protect\usym{2713}). The best results within each category are \textbf{bolded}. See qualitative comparisons in the Supplementary Material.}
\label{tab:surrogate_results}
\resizebox{\columnwidth}{!}{
\setlength{\tabcolsep}{3pt}
\begin{tabular}{@{}lc cccrcr@{}}
\toprule
& & \multicolumn{6}{c}{\textbf{Set A (surrogate)}} \\
\cmidrule(lr){3-8}
\textbf{Method} & \textbf{SI} & \textbf{PSNR} $\uparrow$ & \textbf{RMSE} $\downarrow$ & \textbf{SSIM} $\uparrow$ & $\Delta$\textbf{E} $\downarrow$ & \textbf{LPIPS} $\downarrow$ & \textbf{FID} $\downarrow$ \\
\midrule
CompenNeSt++~\cite{huang2021end} &\usym{2717} & \textbf{23.77} & \textbf{0.066} & \textbf{0.767} & \textbf{5.31} & \textbf{0.281} & \textbf{42.55} \\
CompenHR~\cite{wang2023compenhr} & \usym{2717}& 22.36 & 0.077 & 0.703 & 6.48 & 0.353 & 73.68 \\
DeProCams~\cite{huang2021deprocams} & \usym{2717}& 17.86 & 0.129 & 0.508 & 10.92 & 0.471 & 121.24 \\
DPCS~\cite{li2025dpcs} & \usym{2717}& 16.20 & 0.161 & 0.467 & 12.90 & 0.467 & 123.09 \\
\midrule
FF+CompenNeSt & \usym{2713}& 20.25 & 0.098 & 0.698 & 8.20 & 0.319 & 65.70 \\
FF+CompenNeSt (Origin) & \usym{2713} & 18.57 & 0.119 & 0.571 & 9.28 & 0.350 & 69.76 \\
FF+PANet & \usym{2713} & 19.77 & 0.104 & 0.681 & 8.77 & 0.340 & 76.38 \\
FF+PANet (Origin) & \usym{2713}& 19.05 & 0.112 & 0.635 & 10.00 & 0.380 & 95.38 \\
SIComp (\#surf=1) &\usym{2713} & 21.54 & 0.085 & 0.709 & 7.29 & 0.313 & 57.63 \\
SIComp (\#surf=3) &\usym{2713} & 22.24 & 0.078 & 0.721 & 6.84 & 0.295 & 50.07 \\
SIComp (\#surf=5) & \usym{2713}& \textbf{22.33} & \textbf{0.077} & \textbf{0.722} & \textbf{6.51} & \textbf{0.291} & \textbf{48.41} \\
\bottomrule
\end{tabular}
}
\end{table}

\subsection{Evaluation on surrogate compensation}
We first evaluate performance using the surrogate compensation pipeline~\cite{huang2019end}, which enables a direct comparison between the predicted compensation $\hat{x}$ and the ground-truth input $x$ in a controlled digital environment. Quantitative results for Set A (8 unseen setups and 100 validation images) are detailed in~\cref{tab:surrogate_results}.

As expected, the setup-dependent CompenNeSt++ achieves the highest fidelity, as it is optimized for fixed environments. However, among Setup-Independent (SI) methods, our SIComp (\#surf=5) significantly outperforms existing baselines, including FF+PANet and FF+CompenNeSt, nearly matching the performance of setup-dependent CompenHR. The performance improvement from \#surf=1 to 5 shows the importance of diverse surface priors in modeling complex photometric conditions.

\subsection{Evaluation on real compensation}
\begin{figure*}[tbh]
 \centering
 \includegraphics[width=.9\linewidth]{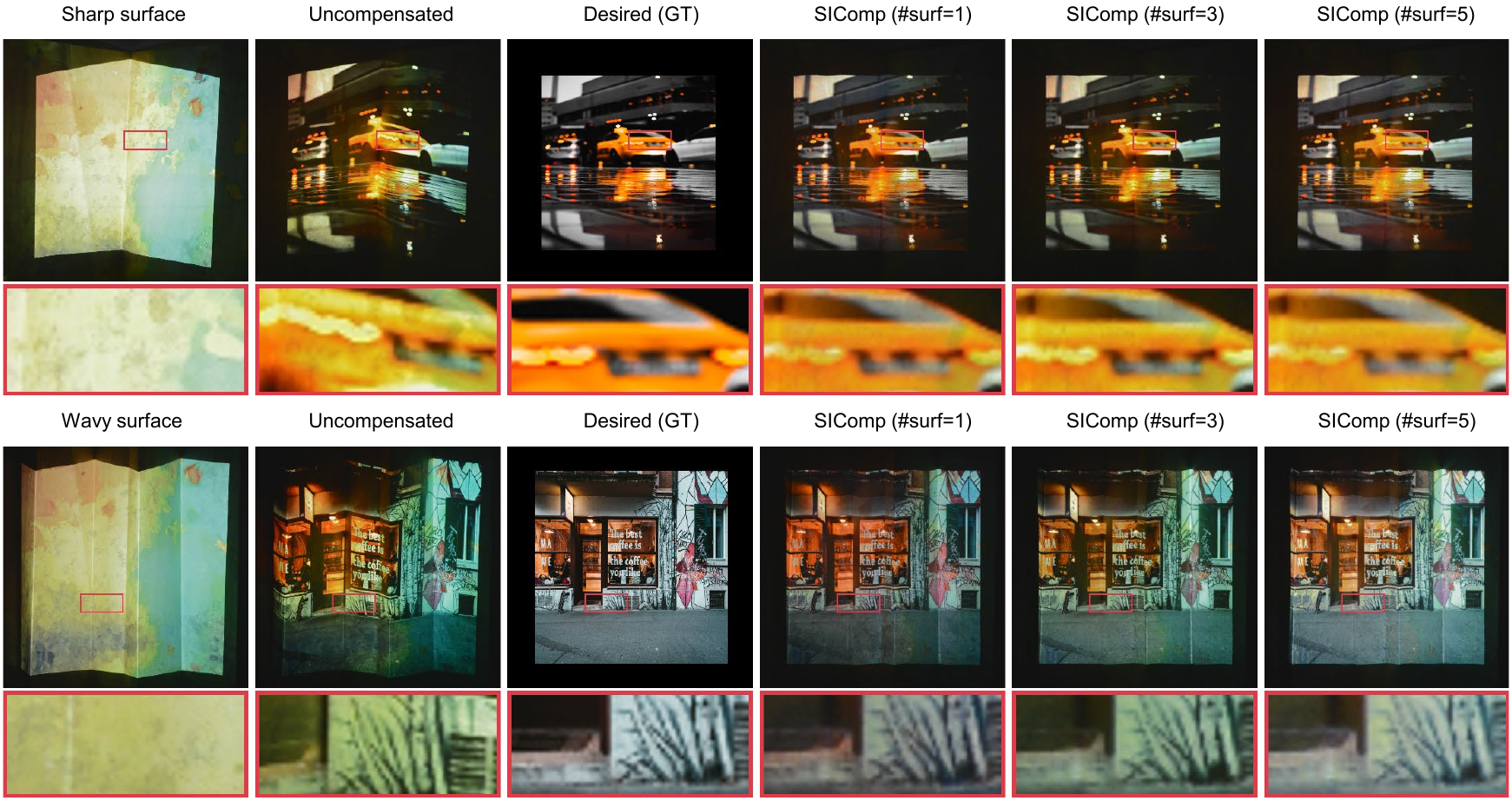}
 \caption{Real compensation of SIComp on \textbf{sharp (top)} and \textbf{wavy (bottom)} surfaces, demonstrating improved quality as \#surf increases from 1 to 5.}
 \label{fig:real_compensation_sharp_wavy}
\end{figure*}

We further validate our approach through physical projection and capture, comparing the camera-captured compensated image against the desired image $x'$. Results for Set A (same devices, unseen setups) and Set B (novel devices, unseen setups) are summarized in~\cref{tab:real_cmp_results} and ~\cref{fig:real_cmp_results}. SIComp results on sharp and wavy surfaces are in~\cref{fig:real_compensation_sharp_wavy}.

\subsubsection{Generalization to unseen setups (Set A)}
SIComp shows the best real compensation quality in the SI group (\cref{tab:real_cmp_results}). While setup-dependent methods~\cite{huang2021end, wang2023compenhr} obtain high fidelity, they require a time-consuming retraining phase ($\sim$1--20 minutes) for every new setup. In contrast, SIComp generalizes to unseen setups without any per-setup training time. As shown in~\cref{tab:efficiency_results}, SIComp achieves a zero-time retraining duration and an inference speed of 12 FPS, which is significantly faster than the retraining-inference cycle of setup-dependent methods. Although SIComp exhibits lower inference speeds compared to the FF-based SI variants, it delivers superior compensation quality. Specifically, SIComp (\#surf=5) outperforms other SI baselines and significantly narrows the performance margin to setup-dependent models. The metric improvement observed as the number of surface priors increases confirms that richer intensity-varying surface priors effectively enhance photometric robustness, making SIComp a better choice  for dynamic, real-world deployment. Note that DeProCams and DPCS suffer from low inference speeds ($<1$ FPS) due to their iterative refinement nature for compensation.

\begin{table}[ht]
\centering
\caption{Efficiency comparison. \textbf{SI} indicates Setup-Independent methods (\protect\usym{2713}). \textbf{BS} indicates the batch size. \textbf{Per Setup} denotes the re-training duration required for each new setup. \textbf{Peak Mem.} denotes the maximum GPU memory during training or fine-tuning. Best results in each category are \textbf{bolded}. See the Supplementary Material for a more detailed efficiency comparison.}
\label{tab:efficiency_results}
\resizebox{\columnwidth}{!}{
\begin{tabular}{@{}l@{\hspace{4pt}}crrrr@{}}
\toprule
& & \multicolumn{2}{c}{\textbf{Train}} & \multicolumn{2}{c}{\textbf{Inference (BS=1)}} \\
\cmidrule(lr){3-4} \cmidrule(lr){5-6}
\textbf{Method} & \textbf{SI} & \textbf{Per Setup} & \textbf{Peak Mem.} & \textbf{Infer Mem.} & \textbf{FPS} $\uparrow$ \\
\midrule
CompenNeSt++~\cite{huang2021end} & \usym{2717} & 20 min 27 s & 23.56 GB & \textbf{0.18 GB} & 5 \\
CompenHR~\cite{wang2023compenhr} & \usym{2717} & \textbf{1 min 16 s} & \textbf{1.22 GB} & 1.05 GB & \textbf{171} \\
DeProCams~\cite{huang2021deprocams} & \usym{2717} & 2 min 01 s & 29.39 GB & 0.63 GB & <1 \\
DPCS~\cite{li2025dpcs} & \usym{2717} & 7 min 13 s & 7.31 GB & 1.83 GB & <1 \\
\midrule
FF+CompenNeSt & \usym{2713} & \textbf{0 s} & 11.26 GB & 0.38 GB & 226 \\
FF+CompenNeSt (Origin) & \usym{2713} & \textbf{0 s} & 7.68 GB & 0.38 GB & 226 \\
FF+PANet & \usym{2713} & \textbf{0 s} & 11.08 GB & \textbf{0.31 GB} & \textbf{296} \\
FF+PANet (Origin) & \usym{2713} & \textbf{0 s} & \textbf{6.55 GB} & \textbf{0.31 GB} & \textbf{296} \\
SIComp (\#surf=1) & \usym{2713} & \textbf{0 s} & 11.83 GB & 0.58 GB & 12 \\
SIComp (\#surf=3) & \usym{2713} & \textbf{0 s} & 11.84 GB & 0.59 GB & 12 \\
SIComp (\#surf=5) & \usym{2713} & \textbf{0 s} & 13.51 GB & 0.61 GB & 12 \\
\bottomrule
\end{tabular}
}
\end{table}

\subsubsection{Generalization to novel devices (Set B)}

To evaluate the generalization capability of SIComp across different devices, we test on \textbf{Set B}, which features entirely novel ProCams devices and unseen setups. As shown in~\cref{tab:real_cmp_results}, SIComp (\#surf=5) achieves competitive results without any fine-tuning. Remarkably, its performance is on par with CompenNeSt++ on several metrics, despite the latter being specifically retrained for these new devices. This confirms that SIComp successfully decouples device-specific characteristics from the compensation logic.

Visual comparisons in~\cref{fig:real_cmp_results} (rows 3-4) also validate these findings. While other SI baselines exhibit edge artifacts or color shifts, SIComp produces high-fidelity results with accurate color restoration across diverse poses and lighting.
SIComp provides a ``plug-and-play'' solution, while maintaining high-quality compensation that is visually indistinguishable from per-setup retrained models.

\begin{figure*}[htb]
 \centering
 \includegraphics[width=.9\linewidth]{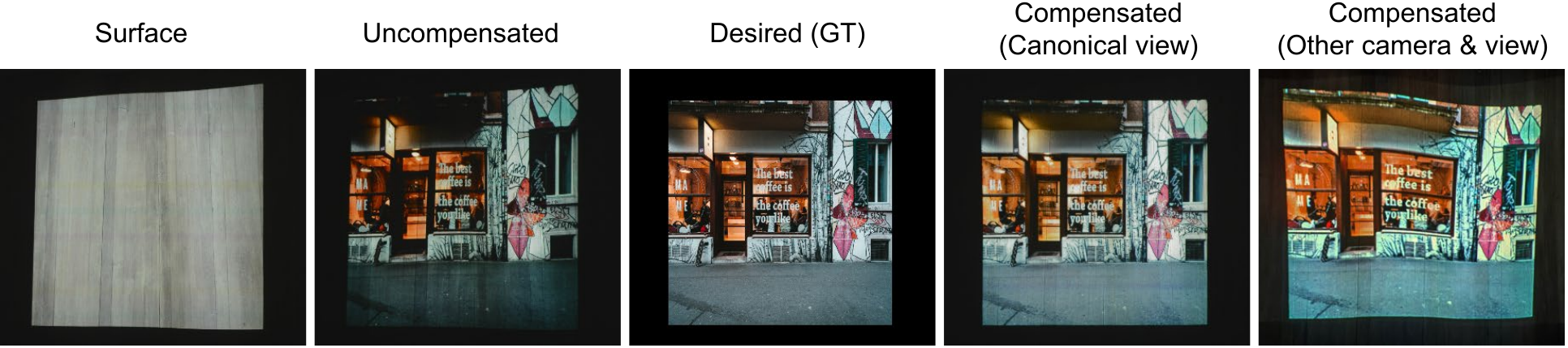}
\caption{View-dependent limitation of SIComp. Our proposed SIComp method is restricted to a single viewpoint. The last two columns display the projected compensations on the surface, captured from two different perspectives: the canonical view (using the same DSLR camera as in columns 1-3) and a side view (using a cell phone camera). The significant degradation in fidelity from the side view clearly demonstrates that view changes compromise compensation quality.}
 \label{fig:compensation_limitations_view_dependent}
\end{figure*}

\subsection{Ablation studies}
We conduct ablation studies on the number of intensity-varying surface priors and loss functions. All  experiments are performed on Set A to ensure controlled and reproducible evaluation. Note that due to space constraints, the detailed analysis of the loss function is provided in the Supplementary Material.

\paragraph{Effect of intensity-varying surface priors.}
\label{effect_num_surf}
To explore the effectiveness of incorporating intensity-varying surface priors, we experiment with different number of surface images (\cref{fig:multi_surface}). Specifically, SIComp (\#surf=1) utilizes one surface image (corresponding to a projection grayscale image level of 64). SIComp (\#surf=3) employs three surface images obtained at projection grayscale image levels 0, 128, and 255. SIComp (\#surf=5) uses five surface images corresponding to projection grayscale image levels 0, 64, 128, 191, and 255.

As detailed in \cref{tab:real_cmp_results} and \cref{fig:real_cmp_results}, increasing the number of surface priors from SIComp (\#surf=1) to SIComp (\#surf=3) leads to a substantial improvement across all key metrics, significantly enhancing compensation performance and robustness. While SIComp (\#surf=5) further improves these results, the gains are relatively marginal compared to the sharp increase observed when moving from \#surf=1 to \#surf=3.

\section{Limitations and Future Work}
While SIComp achieves robust setup-independent performance, it exhibits several constraints that offer directions for future research.

\paragraph{View-dependency.} As shown in~\cref{fig:compensation_limitations_view_dependent}, SIComp assumes a static canonical viewpoint. Significant observer movement introduces degradation due to non-Lambertian surface reflections and projector-camera parallax. Integrating real-time gaze tracking or view-aware appearance modeling remains a promising future direction.

\paragraph{Extreme lighting conditions.} Similar to most projector compensation methods, SIComp faces challenges in extreme scenarios like specularities and occlusions (shadows). Highly reflective surfaces can saturate camera sensors, causing irreversible information loss. While our intensity-varying surface priors enhance robustness, they cannot recover signals in already clipped, overexposed regions. The geometric correction module's dependence on optical flow is also affected by self-occlusions, while the photometric compensation module does not explicitly account for hard shadows. These violations of brightness constancy and flow continuity lead to local artifacts. Extending this framework by incorporating HDR imaging and multi-projectors is an interesting direction to explore in the future.

\section{Conclusion}
In this paper, we introduce SIComp, to our knowledge, the first setup-independent framework for projector compensation. Our framework effectively addresses both geometric and photometric distortions across a wide array of diverse projection setups, encompassing varying surface properties, ambient lighting conditions, and projector-camera poses. A key innovation lies in formulating projector compensation as a generalizable problem, leveraging attention-based modules and incorporating intensity-varying surface priors. This design enables SIComp to adaptively handle unseen configurations without the need for time-consuming retraining or fine-tuning, significantly enhancing the practical applicability of ProCams. Furthermore, we establish a large-scale, diverse dataset to facilitate effective pre-training and promote strong generalization capabilities.

\acknowledgments{
\RaggedRight We thank the anonymous reviewers for their valuable and inspiring comments and suggestions.\par
}
\bibliography{SIComp_arxiv_ref}
\clearpage
\maketitlesupplementary
\appendix  
\setcounter{page}{1}

\vspace*{-1.2cm}
\setcounter{figure}{0}
\renewcommand{\thefigure}{S\arabic{figure}}
\setcounter{table}{0}
\renewcommand{\thetable}{S\arabic{table}}
\setcounter{equation}{0}
\renewcommand{\theequation}{S\arabic{equation}}
\section{Surrogate Compensation Experiment}

\subsection{Qualitative results}
\begin{figure*}[ht]
 \centering
 \includegraphics[width=1\linewidth]{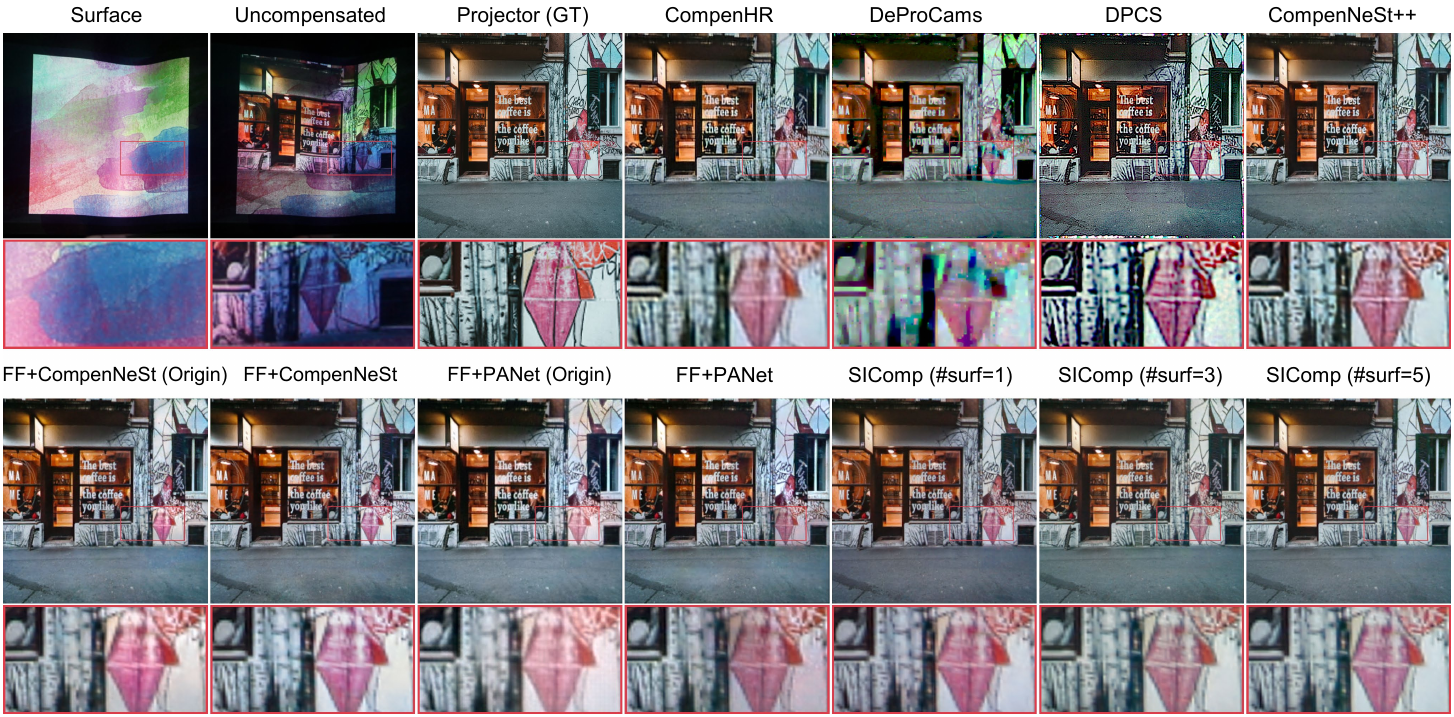}
 \caption{Qualitative results of surrogate experiment. The figure presents results organized into two rows, with seven columns each. The top row of columns presents: the surface, the uncompensated image, the projector (GT), and the compensated results from various setup-dependent methods (CompenHR, DeProCams, DPCS, CompenNeSt++). The bottom row of columns showcases compensated results from setup-independent methods (FF+CompenNeSt, FF+PANet, SIComp (\#surf=1), SIComp (\#surf=3), and SIComp (\#surf=5)). The magnified insets (bottom row, indicated by red boxes) provide detailed visual comparisons.}
 \label{fig:surrogate}
\end{figure*}

Our surrogate experiment qualitative results in~\cref{fig:surrogate} show that setup-independent methods may exhibit minor imperfections, especially at the edges or corners of the projection area due to their generalization nature. However, SIComp (\#surf=5) achieves compensation results that closely resemble, and are often visually comparable to, those of setup-dependent methods like CompenNeSt++~\cite{huang2021end}.

\section{Real Compensation Experiment}
\subsection{Visualizations: same ProCams devices and unseen setups}
\begin{figure*}[t]
 \centering
 \includegraphics[width=1\linewidth]{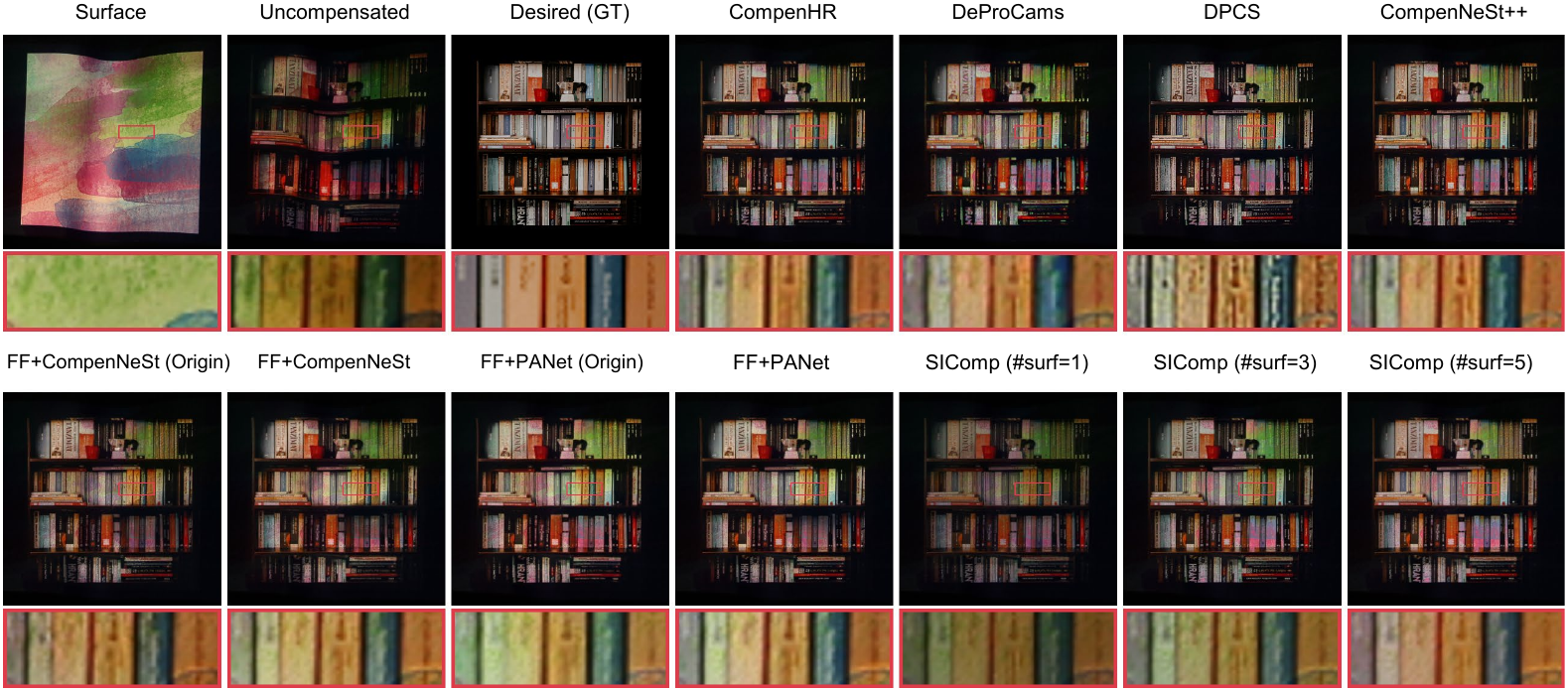}
 \caption{Qualitative compensation results on a setup from Set A, using a Canon camera and an Epson projector. Results are organized into two rows, each with seven columns. The top row of columns presents: the surface, the uncompensated image captured by the camera, the desired (GT), and the compensated results from various setup-dependent methods (CompenHR, DeProCams, DPCS, CompenNeSt++). The bottom row of columns shows compensated results from setup-independent methods (FF+CompenNeSt, FF+PANet, SIComp (\#surf=1), SIComp (\#surf=3), and SIComp (\#surf=5)). The magnified insets (bottom row, indicated by red boxes) provide detailed visual comparisons.}
 \label{fig:additional_real_compensation_1}
\end{figure*}

\begin{figure*}[t]
 \centering
 \includegraphics[width=1\linewidth]{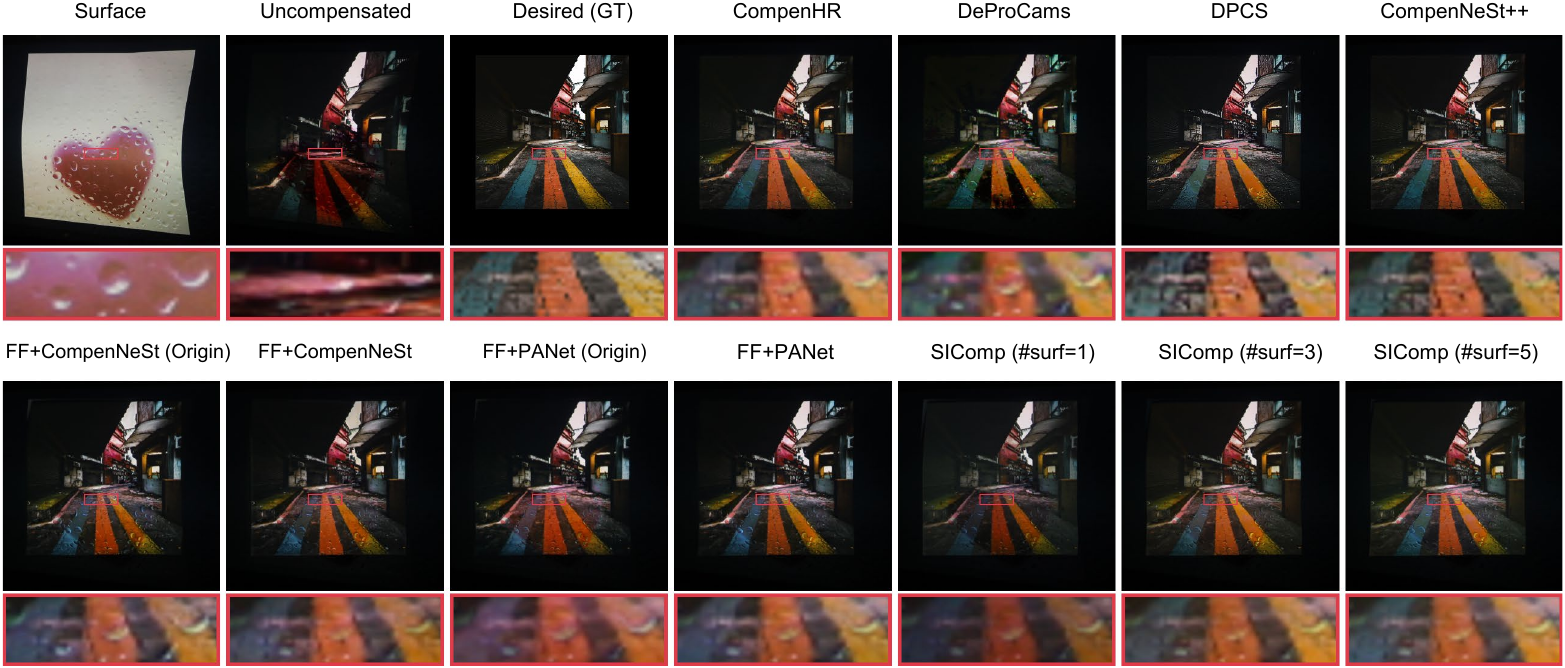}
 \caption{Qualitative compensation results on a setup from Set A, using a Canon camera and an Epson projector. Results are organized into two rows, each with seven columns. The top row of columns presents: the surface, the uncompensated image captured by the camera, the desired (GT), and the compensated results from various setup-dependent methods (CompenHR, DeProCams, DPCS, CompenNeSt++). The bottom row of columns shows compensated results from setup-independent methods (FF+CompenNeSt, FF+PANet, SIComp (\#surf=1), SIComp (\#surf=3), and SIComp (\#surf=5)). The magnified insets (bottom row, indicated by red boxes) provide detailed visual comparisons.}
 \label{fig:additional_real_compensation_3}
\end{figure*}

\begin{figure*}[t]
 \centering
 \includegraphics[width=1\linewidth]{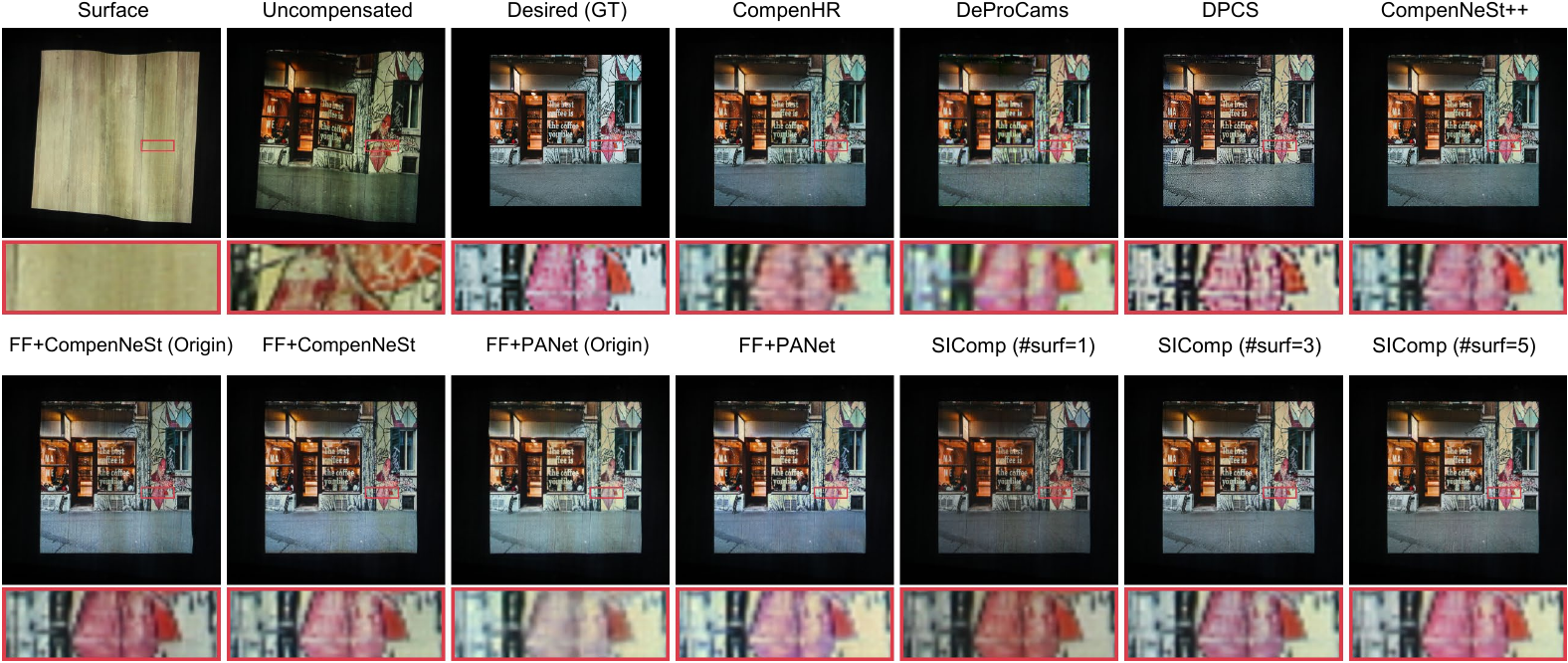}
 \caption{Qualitative compensation results on a setup from Set A, using a Canon camera and an Epson projector. Results are organized into two rows, each with seven columns. The top row of columns presents: the surface, the uncompensated image captured by the camera, the desired (GT), and the compensated results from various setup-dependent methods (CompenHR, DeProCams, DPCS, CompenNeSt++). The bottom row of columns shows compensated results from setup-independent methods (FF+CompenNeSt, FF+PANet, SIComp (\#surf=1), SIComp (\#surf=3), and SIComp (\#surf=5)). The magnified insets (bottom row, indicated by red boxes) provide detailed visual comparisons.}
 \label{fig:additional_real_compensation_4}
\end{figure*}

\begin{figure*}[t]
 \centering
 \includegraphics[width=1\linewidth]{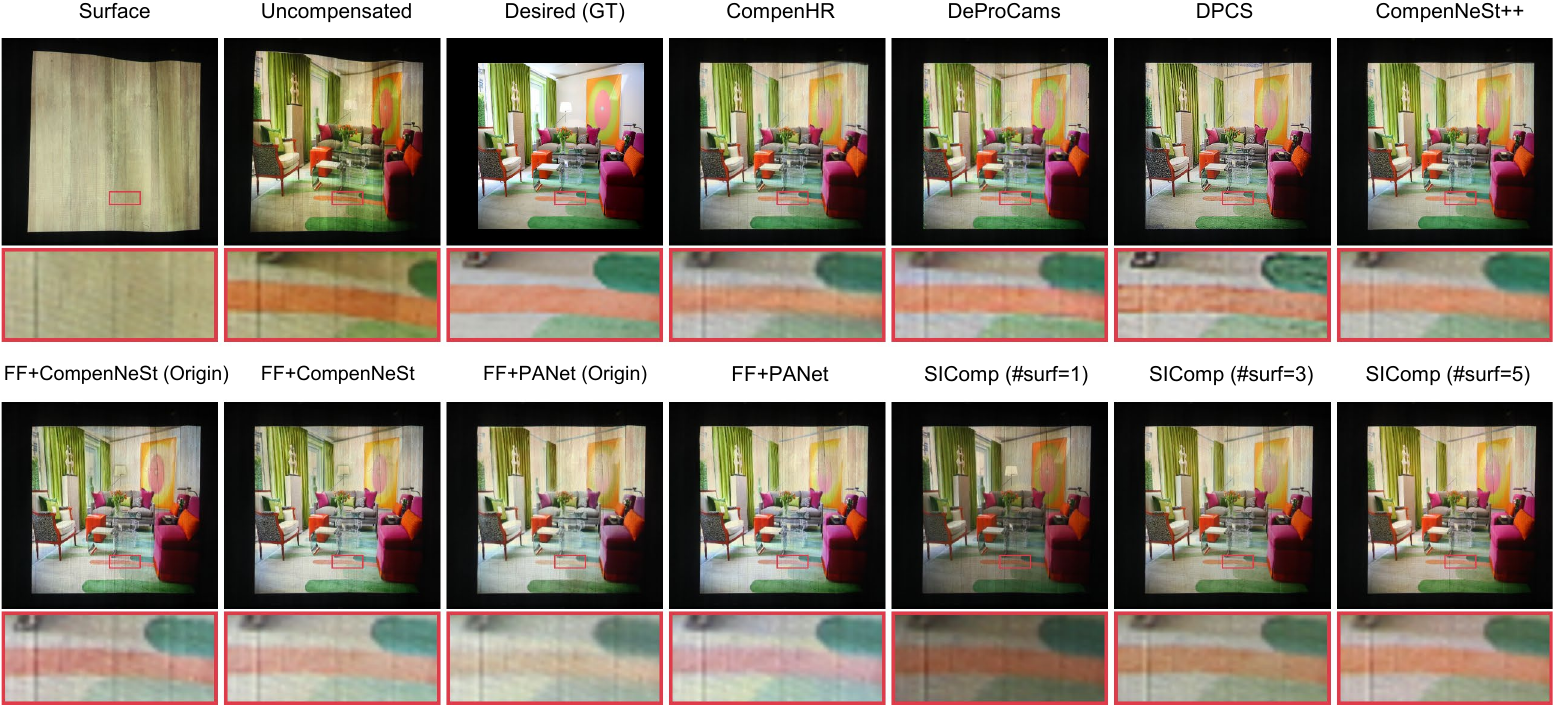}
 \caption{Qualitative compensation results on a setup from Set A, using a Canon camera and an Epson projector. Results are organized into two rows, each with seven columns. The top row of columns presents: the surface, the uncompensated image captured by the camera, the desired (GT), and the compensated results from various setup-dependent methods (CompenHR, DeProCams, DPCS, CompenNeSt++). The bottom row of columns shows compensated results from setup-independent methods (FF+CompenNeSt, FF+PANet, SIComp (\#surf=1), SIComp (\#surf=3), and SIComp (\#surf=5)). The magnified insets (bottom row, indicated by red boxes) provide detailed visual comparisons.}
 \label{fig:additional_real_compensation_5}
\end{figure*}

Complementing the quantitative analysis in Section 4.3.1 of the main manuscript, we provide additional qualitative comparisons across various setups. As illustrated in \cref{fig:additional_real_compensation_1,fig:additional_real_compensation_3,fig:additional_real_compensation_4,fig:additional_real_compensation_5}, SIComp maintains high-fidelity compensation performance without any retraining or fine-tuning.

\subsection{Visualizations: novel ProCams devices and unseen setups}
\begin{figure*}[t]
 \centering
 \includegraphics[width=1\linewidth]{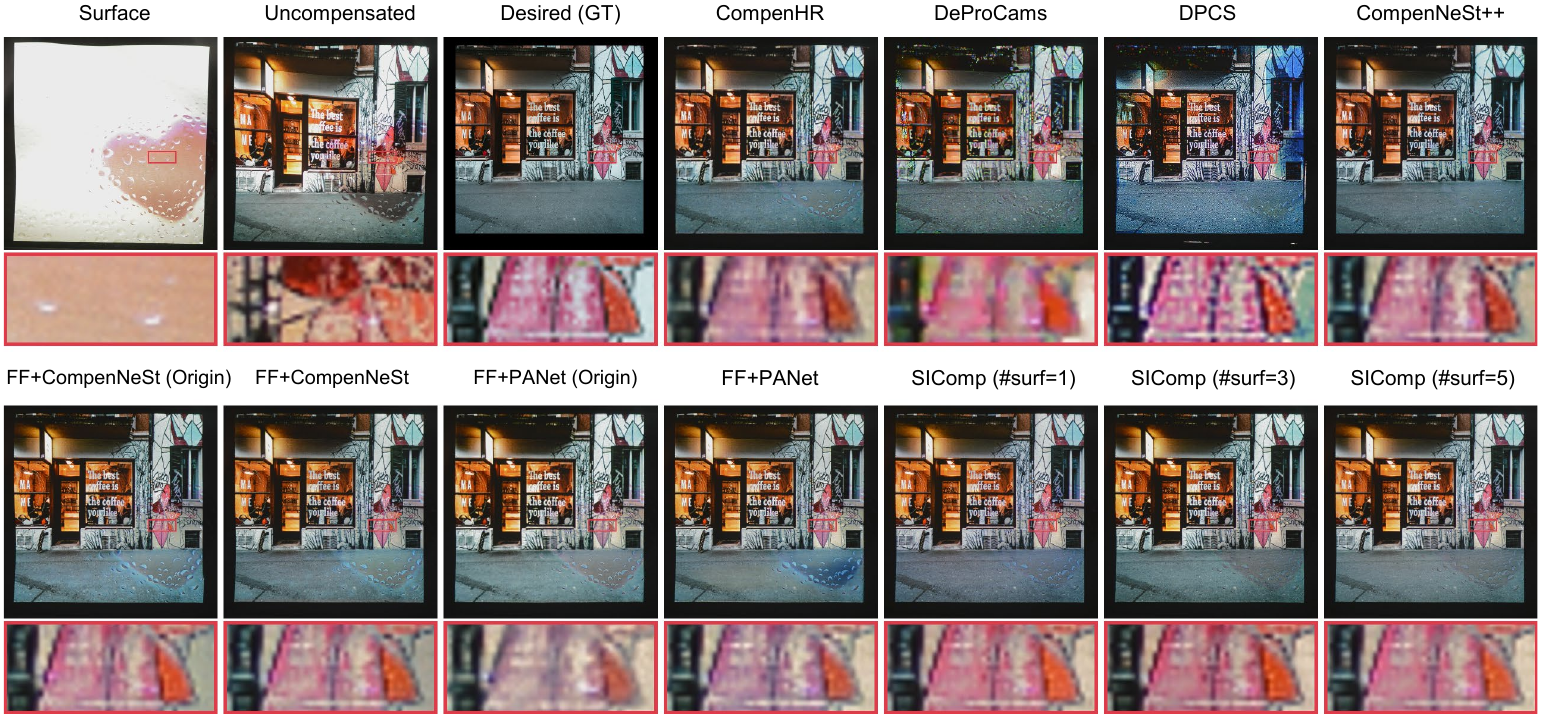}
 \caption{Qualitative compensation results on a setup from Set B, using a Nikon D3200 camera and a Toshiba TDP-T100C projector. Results are organized into two rows, each with seven columns. The top row of columns presents: the surface, the uncompensated image captured by the camera, the desired (GT), and the compensated results from various setup-dependent methods (CompenHR, DeProCams, DPCS, CompenNeSt++). The bottom row of columns shows compensated results from setup-independent methods (FF+CompenNeSt, FF+PANet, SIComp (\#surf=1), SIComp (\#surf=3), and SIComp (\#surf=5)). The magnified insets (bottom row, indicated by red boxes) provide detailed visual comparisons.}
 \label{fig:new_hardware_2}
\end{figure*}

\begin{figure*}[t]
 \centering
 \includegraphics[width=1\linewidth]{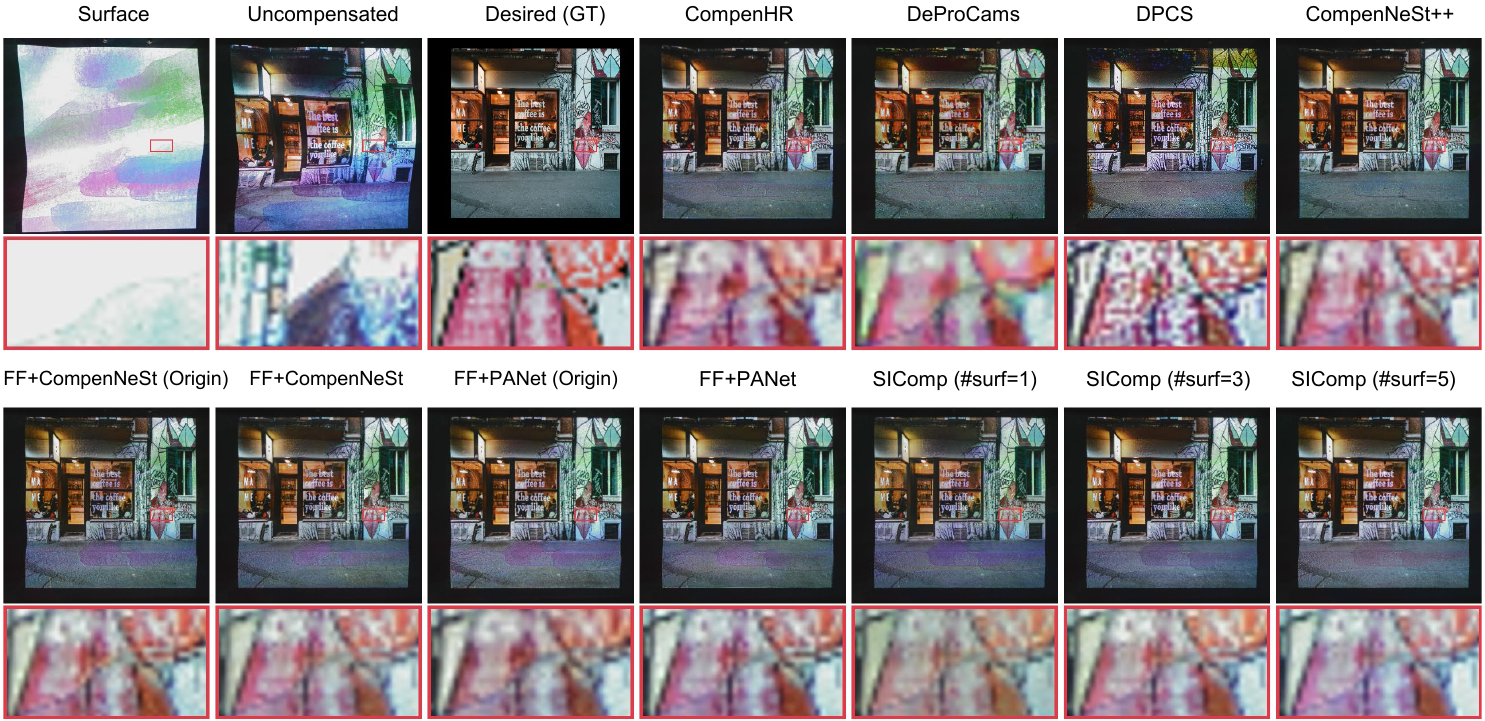}
 \caption{Qualitative compensation results on a setup from Set B, using a Nikon D3200 camera and a Toshiba TDP-T100C projector. Results are organized into two rows, each with seven columns. The top row of columns presents: the surface, the uncompensated image captured by the camera, the desired (GT), and the compensated results from various setup-dependent methods (CompenHR, DeProCams, DPCS, CompenNeSt++). The bottom row of columns shows compensated results from setup-independent methods (FF+CompenNeSt, FF+PANet, SIComp (\#surf=1), SIComp (\#surf=3), and SIComp (\#surf=5)). The magnified insets (bottom row, indicated by red boxes) provide detailed visual comparisons.}
 \label{fig:new_hardware_3}
\end{figure*}

Complementing the quantitative analysis in Section 4.3.2 of the main manuscript, we provide additional visual results on the Set B dataset.  As shown in~\cref{fig:new_hardware_3,fig:new_hardware_2}, SIComp achieves high-fidelity compensation and geometric alignment across various camera settings (e.g., ISO, shutter speed) and surface geometries within this dataset. These results were also obtained without further fine-tuning, demonstrating the model's ability to handle hardware characteristics that differ from the training setup. These visual examples provide qualitative evidence that SIComp can maintain robust performance when applied to this novel device configuration.

\section{Ablation Study on Loss Functions}
To investigate the optimal combination of loss terms for balancing pixel-level accuracy and perceptual quality in SIComp's training, we compare four distinct loss configurations. (1) $\ell_1$ + $\text{SSIM}$, (2) $\ell_1$ + $\text{SSIM}$ + $\Delta$E, (3) $\ell_1$ + $\text{SSIM}$ + $\text{LPIPS}$, and (4) $\ell_1$ + $\text{SSIM}$ + $\Delta$E + $\text{LPIPS}$. The inclusion of perceptual losses, $\Delta$E and LPIPS, is hypothesized to improve visual quality by aligning network predictions more closely with human perception.

As detailed in~\cref{tab:loss_functions}, our experiments reveal distinct trade-offs among different loss configurations. The simplest combination, $\ell_1 + \text{SSIM}$, achieves the best performance on traditional pixel-level metrics, securing the highest PSNR, lowest RMSE, and highest SSIM.

However, when perceptual losses are introduced, the model's performance exhibits a clear shift towards perceptual fidelity, though not without trade-offs. Specifically, the $\ell_1$ + $\text{SSIM}$ + $\Delta$E combination emerges as the strongest performer in perceptual metrics, achieving the lowest $\Delta$E. This confirms the hypothesis that carefully chosen perceptual losses can indeed improve visual quality by better matching human perception. Noteably, this perceptual enhancement comes with a slight compromise in strict pixel-level accuracy (e.g., slightly higher RMSE and lower PSNR/SSIM) when compared to the $\ell_1 + \text{SSIM}$ baseline.

\section{Efficiency and Resource Consumption}
~\cref{tab:efficiency_final_formatted} reports the detailed comparison of efficiency and resource consumption between \textbf{setup-dependent (SD)} and \textbf{setup-independent (SI)} models. A fundamental advantage of the SI approach is its one-time training paradigm. As shown in~\cref{tab:efficiency_final_formatted}, SD models (e.g., CompenNeSt++) require a complete retraining process (approximately 20 minutes) every time the setup changes. In contrast, SI models require only a single training session encompassing both pre-training and fine-tuning. Once this process is completed, the models can be directly deployed to new setups without further training or fine-tuning. This eliminates repetitive training overhead, making the SI models significantly more flexible for practical deployment.

\begin{table}[ht]
  \centering
  \caption{
  Comparison of different loss functions. Results are averaged over the 8 setups in Set A.
  }
  \label{tab:loss_functions}
    \begin{tabular}{l @{\hspace{0.5em}} c @{\hspace{0.4em}} c @{\hspace{0.4em}} c @{\hspace{0.4em}} c @{\hspace{0.4em}} c @{\hspace{0.4em}} c @{\hspace{0.5em}}}
      \toprule
      \textbf{Loss} & \textbf{PSNR} $\uparrow$  & \textbf{RMSE} $\downarrow$ & \textbf{SSIM} $\uparrow$ & $\Delta$\textbf{E} $\downarrow$ & \textbf{LPIPS} $\downarrow$ & \textbf{FID} $\downarrow$  \\
      \midrule
      $\ell_1$+SSIM & \textbf{22.3325} & \textbf{0.0774} & \textbf{0.7222} & 6.5113 & 0.2909 & 48.4108 \\
      $\ell_1$+SSIM+$\Delta$E & 22.2989 & 0.0777 & 0.7190 & \textbf{6.3698} & 0.2896 & 47.8011 \\
      $\ell_1$+SSIM+LPIPS & 22.2762 & 0.0779 & 0.7159 & 6.3944 & 0.2836 & 46.3983 \\
      $\ell_1$+SSIM+$\Delta$E+LPIPS & 22.2950 & 0.0778 & 0.7158 & 6.3852 & \textbf{0.2826} & \textbf{45.7490} \\
      \bottomrule
    \end{tabular}
\end{table}

\begin{table*}[t]
\centering
\caption{Efficiency and resource consumption statistics. All input and output image resolutions are $(600 \times 600)$. We provide a comparative analysis between \textbf{setup-dependent (SD)} models and \textbf{setup-independent (SI)} models. SI methods are marked with $(\protect\usym{2713})$. The training efficiency is evaluated via \textbf{SD Train Time} for SD models, and \textbf{SI Pre-train/Fine-tuning Time} for SI models. For a fair comparison, all inference metrics are normalized to a single-stream execution with a batch size of 1 (\textbf{BS}=1). N/A indicates that the corresponding process is not applicable to the method. \textbf{Flow Time} represents the constant temporal cost (0.75 s) for flow estimation.}

\label{tab:efficiency_final_formatted}
\resizebox{\linewidth}{!}{
\begin{tabular}{lcccccccccc}
\toprule
\textbf{Method} & \textbf{SI} & \textbf{SD Train} & \textbf{SI Pre-train} & \textbf{SI Fine-tuning} & \textbf{Train} & \textbf{Train Peak} & \textbf{Infer.} & \textbf{Infer.} & \textbf{Flow} & \textbf{Infer.} \\
& & \textbf{Time} & \textbf{Time} & \textbf{Time} & \textbf{BS} & \textbf{Mem} & \textbf{Mem} & \textbf{BS} & \textbf{Time} & \textbf{FPS} \\
\midrule
CompenNeSt++~\cite{huang2021end} & \usym{2717} & 20 min 27 s & N/A & N/A & 48 & 23.56 GB & 0.18 GB & 1 & N/A & 5 \\
CompenHR~\cite{wang2023compenhr} & \usym{2717} & 1 min 16 s & N/A & N/A & 4 & 1.22 GB & 1.05 GB & 1 & N/A & 171 \\
DeProCams~\cite{huang2021deprocams} & \usym{2717} & 2 min 01 s & N/A & N/A & 24 & 29.39 GB & 0.63 GB & 1 & N/A & $<$ 1 \\
DPCS~\cite{li2025dpcs} & \usym{2717} & 7 min 13 s & N/A & N/A & 4 & 7.31 GB & 1.83 GB & 1 & N/A & $<$ 1 \\
\midrule
FF+CompenNeSt & \usym{2713} & 0 s & 131 min 32 s & 100 min 58 s & 6 & 11.26 GB & 0.38 GB & 1 & 0.75 s & 226 \\
FF+CompenNest (Origin) & \usym{2713} & 0 s & N/A & 141 min 20 s & 4 & 7.68 GB & 0.38 GB & 1 & 0.75 s & 226 \\
FF+PANet & \usym{2713} & 0 s & 75 min 28 s & 101 min 43 s & 6 & 11.08 GB & 0.31 GB & 1 & 0.75 s & 296 \\
FF+PANet (Origin) & \usym{2713} & 0 s & N/A & 12 min 34 s & 4 & 6.55 GB & 0.31 GB & 1 & 0.75 s & 296 \\
SIComp (\#surf=1) & \usym{2713} & 0 s & 454 min 49 s & 116 min 20 s & 6 & 11.83 GB & 0.58 GB & 1 & 0.75 s & 12 \\
SIComp (\#surf=3) & \usym{2713} & 0 s & 450 min 46 s & 117 min 32 s & 6 & 11.84 GB & 0.59 GB & 1 & 0.75 s & 12 \\
SIComp (\#surf=5) & \usym{2713} & 0 s & 463 min 47 s & 126 min 10 s & 6 & 13.51 GB & 0.61 GB & 1 & 0.75 s & 12 \\
\bottomrule
\end{tabular}
}
\end{table*}

\end{document}